\documentclass[runningheads]{llncs}
\usepackage{tabularx}
\usepackage{ctable}
\usepackage[position=top]{subfig}
\usepackage{graphicx}
\usepackage{multirow}
\newcolumntype{P}[1]{>{\centering\arraybackslash}p{#1}}

\usepackage[colorlinks=true,allcolors=blue]{hyperref}

\usepackage[bottom]{footmisc}
\usepackage{float} % -------------------------------- 
\begin{document}
\title{ Weakly-Supervised Segmentation for Disease Localization in Chest X-Ray Images}
% \thanks{Supported by organization x.}
\titlerunning{Weakly-Supervised Segmentation in Chest X-Ray images}
% If the paper title is too long for the running head, you can set
% an abbreviated paper title here
%
% \newcounter{savecntr}
\author{Ostap Viniavskyi\thanks{These authors contributed equally  to
the work}\inst{1}
\and Mariia Dobko$^*$\inst{1,2} 
\and
Oles Dobosevych \inst{1}}
\authorrunning{Viniavskyi et al.}
% First names are abbreviated in the running head.
% If there are more than two authors, 'et al.' is used.
%
\institute{The Machine Learning Lab at Ukrainian Catholic University, Lviv, Ukraine 
% \url{https://apps.ucu.edu.ua/en/mllab/} \\
\email{\{viniavskyi,dobko\_m,dobosevych\}@ucu.edu.ua} \and
SoftServe, Ukraine \\
% \url{http://https://www.softserveinc.com/en-us} \\
\email{\{mdobk\}@softserveinc.com}}
\maketitle              % typeset the header of the contribution
\begin{abstract}
Deep Convolutional Neural Networks have proven effective in solving the task of semantic segmentation. However, their efficiency heavily relies on the pixel-level annotations that are expensive to get and often require domain expertise, especially in medical imaging. Weakly supervised semantic segmentation helps to overcome these issues and also provides explainable deep learning models.
\par
In this paper, we propose a novel approach to the semantic segmentation of medical chest X-ray images with only image-level class labels as supervision. We improve the disease localization accuracy by combining three approaches as consecutive steps. First, we generate pseudo segmentation labels of abnormal regions in the training images through a supervised classification model enhanced with a regularization procedure. The obtained activation maps are then post-processed and propagated into a second classification model—Inter-pixel Relation Network, which improves the boundaries between different object classes. Finally, the resulting pseudo-labels are used to train a proposed fully supervised segmentation model.
\par
We analyze the robustness of the presented method and test its performance on two distinct datasets: PASCAL VOC 2012 and SIIM-ACR Pneumothorax. We achieve significant results in the segmentation on both datasets using only image-level annotations. We show that this approach is applicable to chest X-rays for detecting an anomalous volume of air in the pleural space between the lung and the chest wall. Our code has been made publicly available\footnote{Implementation is available at \url{https://github.com/ucuapps/WSMIS}}.

\keywords{Weakly-supervised learning  \and Segmentation \and Deep Learning \and Chest X-rays \and Disease localization \and Explainable models}
\end{abstract} 

% \newcommand\blfootnote[1]{%
%   \begingroup
%   \renewcommand\thefootnote{}\footnote{#1}%
%   \addtocounter{footnote}{-1}%
%   \endgroup
% }
% \blfootnote{Our code is available at \url{https://github.com/ucuapps/WSMIS}} 

\section{Introduction}

Applications of Convolutional Neural Networks to medical images have recently produced efficient solutions for a vast variety of medical problems, such as segmentation of lung nodules in computed tomography (CT) scans \cite{gruetzemacher20183d}, lesions detection in mammography images \cite{abdelhafiz2019deep}, segmentation of brain gliomas from MRI images \cite{archa2018segmentation} and others \cite{ronneberger2015u,skourt2018lung}.
One of the greatest challenges for using deep learning methods in medicine is the lack of large annotated datasets, especially with pixel-level labeled data.
% In order to create models which can effectively segment and detect disease in medical images, the immense datasets of labeled data are required. 
Creating such datasets is often very expensive and time consuming. For instance, Lin et al. \cite{lin2014microsoft} calculated that collecting bounding boxes for each class is about 15 times faster than producing a ground-truth pixel-wise segmentation mask; getting image-level labels is even easier. Moreover, domain expertise is required to label medical data, which poses another challenge as the doctor's time is costly and could more effectively be used for a patient's diagnosis and disease treatment. Working with image-level annotations also decreases the probability of disagreement between experts, since pixel-wise annotations tend to have more noise and vary among labelers. 
\par
To decrease the resources spent on labeling while preserving its quality, we propose a novel weakly-supervised approach to image segmentation that uses only image-level labels. Our method is domain-independent; we have tested it on several distant datasets, including popular PASCAL VOC 2012 \cite{pascal-voc-2012}, and a medical dataset, SIIM-ACR Pneumothorax \cite{siimpneumothorax}. We achieve 64.6 mean intersection-over union (mIoU) score on PASCAL VOC 2012 \cite{pascal-voc-2012} validation set. Our method is capable of segmenting medical images with limited supervision achieving 76.77 mIoU score on the test set of SIIM-ACR Pneumothorax dataset \cite{siimpneumothorax}. The automatic approach to finding Pneumothorax could be used to triage chest radiographs requiring priority interpretation, to rapidly identify critical cases, and also provide a second-opinion for radiologists to make a more confident disease diagnosis.

\section{Related Work}
The objective of weakly-supervised segmentation is to create models capable of pixel-wise segmentation based on image-level labels. The existing approaches can be categorized by their methodologies into four groups: Expectation-Ma\-xi\-mi\-za\-tion, Multiple Instance Learning, Self-Supervised Learning, and Object Proposal Class Inference \cite{chan2019comprehensive}. In this paper, we follow the self-supervised paradigm, which suggests training a fully supervised segmentation model on the created pseudo-pixel-level annotations, also known as Class Activation Maps (CAM) \cite{zhou2016learning}, which are extracted from the classification network. 
This paradigm is the most challenging one as it leads to the least informative form of weak supervision providing no location information for the objects. However, judging from the quantitative performance on PASCAL VOC 2012 \cite{pascal-voc-2012} validation set, the top five methods of weakly-supervised segmentation use the self-supervised learning approach \cite{chan2019comprehensive}.
\par
Many methods of self-supervised learning for semantic segmentation have been recently suggested. Kolesnikov et al. \cite{kolesnikov2016seed} propose Seed Expand Constrain (SEC) method, which trains a CNN, applies CAM to produce pseudo-ground-truth segments, and then trains a Fully Convolutional Network (FCN) optimizing three losses: one for the generated seeds, another for the image-level label, and, finally, a constraint loss against the maps processed by Conditional Random Fields (CRF). Huang et al. \cite{huang2018weakly} introduce Deep Seeded Region Growing (DSRG), which propagates class activations from high-confidence regions to adjacent regions with a similar visual appearance by applying a region-growing algorithm on the generated CAM. Lee et al. \cite{lee2019ficklenet} present FickleNet, which trains a CNN at the image level with a regularization step represented as a center-fixed spatial dropout in the later convolutional layers, and then runs Grad-CAM \cite{selvaraju2017grad} multiple times to generate a thresholded pseudo-labels for a segmentation step. Another approach, proposed by Ahn et al. \cite{ahn2019weakly}, suggests using IRNet \cite{ahn2019weakly}, which takes the random walk from low-displacement field centroids in the CAM up until the class boundaries as the pseudo-ground-truths for training an FCN. Ahn et al. \cite{ahn2019weakly} focus on the segmentation of the individual instances estimating two types of features in addition to CAM: a class-agnostic instance map and pairwise semantic affinities.
\par
The weakly-supervised semantic segmentation on medical datasets has been explored in \cite{lu2020weakly,agarwal2020weakly,demiray2019weakly,cai2018accurate,qu2019weakly}. On the other hand, Ouyang et al. \cite{ouyang2019weakly} combine the weakly-annotated data with well-annotated cases to segment Pneumothorax in chest X-rays. In our approach, we do not use any form of supervision besides image-level labels. We focus on developing a standardized method, which is efficient for various data types, especially for medical images.

\section{Methodology}
Our method can be split into three consecutive steps: Class Activation Maps generation, map enhancement with Inter-pixel Relation Network, and segmentation. After each step, we also add one or more post-processing techniques such as CRF, thresholding, noise filtering (small regions with low confidence). 

% \subsection{CAM generation}
\par
\subsubsection{Step 1. CAM generation} First, we train fully-supervised classification models on image-level labels. The two tested architectures for this step were ResNet50 \cite{he2016deep} and VGG16 \cite{simonyan2014very} with additional three convolutional layers followed by ReLU activation. We also replace stride with dilation \cite{yu2015multi} in the last convolutional layers to increase the size of the final feature map while decreasing the output stride from 32 to 8. We improve the classification performance by including a regularization term, inspired by FickleNet \cite{lee2019ficklenet}. For this, we use DropBlock \cite{ghiasi2018dropblock}---a dropout technique, which to our best knowledge has not been tried in previous works on weakly-supervised segmentation.  The trained models are then used to retrieve activation maps by applying the Grad-CAM++ \cite{chattopadhay2018grad} method. The resulting maps serve as pseudo labels for segmentation task. 

% \subsection{IRNet model}
\par
\subsubsection{Step 2. IRNet}On the second step, IRNet \cite{ahn2019weakly}  takes the generated CAM and trains two output branches that predict a displacement vector field and a class boundary map, correspondingly. They take feature maps from all the five levels of the same shared ResNet50 \cite{he2016deep} backbone. The main advantage of IRNet \cite{ahn2019weakly} is its ability to improve boundaries between different object classes. We train it on the generated maps, thus no extra supervision is required. This step allows us to obtain better pseudo-labels before proceeding to segmentation. To our best knowledge, this approach has not been used in the medical imaging domain before.  

\par
% \subsection{Segmentation network}
\subsubsection{Step 3. Segmentation} For the segmentation step, we train DeepLabv3+ \cite{chen2018encoder} and U-Net \cite{ronneberger2015u} models with different backbones, which have proven to produce reliable results in fully supervised semantic segmentation on medical images \cite{skourt2018lung,ronneberger2015u}. The used backbones include ResNet50 \cite{he2016deep} and SEResNeXt50 \cite{hu2018squeeze}. We modify the binary cross-entropy (BCE) loss during segmentation by adding weights to a positive class to prevent overfitting towards normal cases.

% TODO: subsubsections

\section{Experiments and Results}

\subsection{Reproducibility}
\label{repro}
PyTorch \cite{NEURIPS2019_9015} was used for implementing and training all steps of our approach: extracting localization maps via the classification networks, improving obtained maps with IRNet \cite{ahn2019weakly} and segmenting the image during segmentation task. All the experiments were performed on four Nvidia Tesla K80 GPUs. 

\subsection{Datasets and evaluation metric}
We conduct experiments on two datasets: PASCAL VOC 2012 \cite{pascal-voc-2012} and  SIIM-ACR Pneumothorax \cite{siimpneumothorax}. We evaluate the quality of our pseudo-ground-truth and the performance of the segmentation model trained on them using mIoU. 
\par
PASCAL VOC 2012 \cite{pascal-voc-2012} is an image segmentation benchmark dataset containing 20 object classes, and a background class. As in other works on weakly-supervised segmentation, we train our models using augmented 10,582 training images with image-level labels. We report mIoU for 1,449 validation images.
%and 1,456 test images. The results for the test images were obtained on the official PASCAL VOC evaluation server.

\par
SIIM-ACR Pneumothorax \cite{siimpneumothorax} is a competition that provides an open dataset of chest X-ray images with pixel-wise annotation for regions affected by Pneumothorax: a collapsed lung, where an abnormal volume of air is formed in the pleural space between the lung and the chest wall. This dataset was formed from a subset of ChestX-ray14 dataset \cite{wang2017chestx}, but relabeled by professional radiologists, and additionally annotated on a pixel level. The specified competition has two stages; ground truth labels are provided only for the first, the second is evaluated on the competition website. Thus, we divided images from the first stage into three sets: train, validation and test. Totally, 12,047 frontal-view chest X-ray cases are in the dataset. We use 2,379 positive and 8,296 negative images for training, 145 and 541 for validation, 145 and 541 for the test.  
% https://www.kaggle.com/c/siim-acr-pneumothorax-segmentation/discussion/100383
% https://www.kaggle.com/c/siim-acr-pneumothorax-segmentation/discussion/100398

\subsection{Data challenges}
\label{data}
SIIM-ACR Pneumothorax \cite{siimpneumothorax} dataset has a severe class imbalance problem. The number of normal cases exceeds approximately 4 times the number of positive ones. In order to prevent overfitting towards healthy patients we use various augmentation techniques such as scaling, rotation, blur, brightness adjustment, and horizontal flipping. We also add sampling in our data loader during training, which selects the constant ratio between negative and positive class. Another challenge in this dataset is the size of regions of interest. Pneumothorax usually affects a very small area of lungs resulting in a high disbalance among the image pixels. We solve this problem by adding weights for positive class to binary cross-entropy loss. Due to these data challenges we evaluate the performance of our method on SIIM-ACR Pneumothorax not only for all images in validation and test sets, but also separately for positive cases, see Table \ref{siimeval}.

\begin{table}[b!]
\centering
\caption{Influence of DropBlock on PASCAL VOC 2012. Comparison of the classification model trained without regularization to a model with DropBlock.}
\vspace{0.5em}
\begin{tabular}{l|P{3.5cm}}
\hline
Model & Multilabel F1-score (\%) \\
\specialrule{.1em}{.05em}{.05em} 
ResNet50 & 88.08 \\
\hline
ResNet50 with DropBlock regularization & 88.2  \\
\hline
\end{tabular}
\label{dropblock}
% \vspace{-2.5em}
\end{table}

\begin{table}[b!]
\centering
\caption{Comparison of CAM generation techniques on PASCAL VOC 2012.}
\vspace{0.5em}
\begin{tabular}{l|l|l|l}
\hline
Classification model & CAM extraction method & mIoU train & mIoU val \\
\specialrule{.1em}{.05em}{.05em} 
VGG16 & Cam-Grad & 0.4137 & 0.3511\\
\hline
VGG16 & Cam-Grad++ & \bfseries 0.4176 & \bfseries 0.3941 \\
\hline
\end{tabular}
\label{pascalcam}
\end{table}

\subsection{Experiments}
\subsubsection{Step 1. CAM generation}
For classification we implement ResNet50, and VGG16. As suggested in previous work \cite{jiang2019integral}, we added three convolutional layers on the top of the fully-convolutional backbone, each of which is followed by a ReLU. The conducted experiments show that adding DropBlock regularization to our classification models improves their performance, see Table \ref{dropblock}. 
For both datasets, the best classification results were achieved using VGG16, which was, thus, selected as the final model for this task. We test two methods for generating pseudo-annotations: Grad-CAM \cite{selvaraju2017grad}, and Grad-CAM++ \cite{chattopadhay2018grad}. Our experiments show that Grad-CAM++ \cite{chattopadhay2018grad}, which utilizes a regularization that Grad-CAM is lacking of,  provides better object localization through visual explanations of model predictions; cf. Table \ref{pascalcam}.

\subsubsection{Step 2. IRNet}
For both datasets as post-processing of maps produced at Step 1, we use thresholding and then refine the pseudo-maps by dense CRF to better capture object shapes. The resulting annotations are used to train IRNet. 

\subsubsection{Step 3. Segmentation}
The obtained maps after IRNet step are used as the pseudo-labels for segmentation. We implement three networks to complete this task: U-Net \cite{ronneberger2015u}, DeepLabv3 \cite{chen2017rethinking}, and DeepLabv3+ \cite{chen2018encoder}. We report results on PASCAL VOC 2012 produced by DeepLabv3+, as it shows better performance than DeepLabv3 with the same ResNet50 backbone. For SIIM-ACR Pneumothorax, however, U-Net with SEResNeXt50 \cite{hu2018squeeze} backbone shows the best results.

\subsection{Training and optimization}
\label{optim}
For all the models, three optimization methods are examined: SGD, Adam and RAdam \cite{liu2019variance}. For Pneumothorax segmentation, SGD optimizer is applied with the learning rate initiated as 6e-5 and gradually decreasing each epoch, whereas momentum is set to 0.9, and weight decay to 1e-6. The size of network inputs is 512x512, the batch is 48, and it is balanced according to class distribution using augmentations to increase the sample of positive cases.

\subsection{Results}
\par
The comparison of segmentation methods using the same chest X-ray datasets is not simple due to the challenge of finding public medical data. Moreover, this work is the first to present the results of weakly-supervised segmentation methods on SIIM-ACR Pneumothorax \cite{siimpneumothorax} data. However, the performance of our models is comparable to Ouyang et al. \cite{ouyang2019weakly}, who reported their scores on a collected closed dataset of Pneumothorax. These authors train their method with different combinations of well- and weakly-annotated data, whereas our method uses only image-level labels. In Table \ref{siimeval} we show how the results improve with each step of our approach; the result of Ouyang et al. \cite{ouyang2019weakly} model trained on 400 weakly-annotated and 400 well-annotated cases is specified too. 

\begin{table}[h!]
\vspace{-1.5em}
\centering
\caption{Results on SIIM-ACR Pneumothorax validation and test sets after each step of our method. Calculated for only positive cases (pos.), and for the whole set, including the healthy patients (all). The Ouyang et al. method, whose result is demonstrated, was trained on 400 weakly-annotated and 400 well-annotated cases.}
\vspace{1.0em}
\begin{tabularx}{\textwidth}{l|l|P{1.5cm}|P{1.5cm}|P{1.5cm}|P{1.5cm}}
%\cline{3-6}
\hline
 \multicolumn{2}{c}{}  &  \multicolumn{2}{|c}{mIoU val}  &  \multicolumn{2}{|c}{mIoU test} \\
\hline
Dataset & Method &  pos. & all &  pos. & all \\
\specialrule{.1em}{.05em}{.05em} 
SIIM-ACR Pneum. \cite{siimpneumothorax} &Step 1. CAM & 0.117 & 0.7633  & 0.142 & 0.7590\\
\hline
SIIM-ACR Pneum. \cite{siimpneumothorax} &Step 2. IRNet &  0.122 & 0.7645 & 0.154 &    0.7607 \\
\hline
SIIM-ACR Pneum. \cite{siimpneumothorax} &Step 3. Segm. & 0.148 & 0.7649 & 0.162 & 0.7677\\
\specialrule{.1em}{.05em}{.05em} 
Custom \cite{ouyang2019weakly} & Ouyang et al.\cite{ouyang2019weakly} & -  & - & - & 0.669 \\
\hline
\end{tabularx}
\label{siimeval}
\end{table}

\begin{figure}[t!]
  \centering % <-- added
  \subfloat[Image] {
    \includegraphics[width=0.18\linewidth]{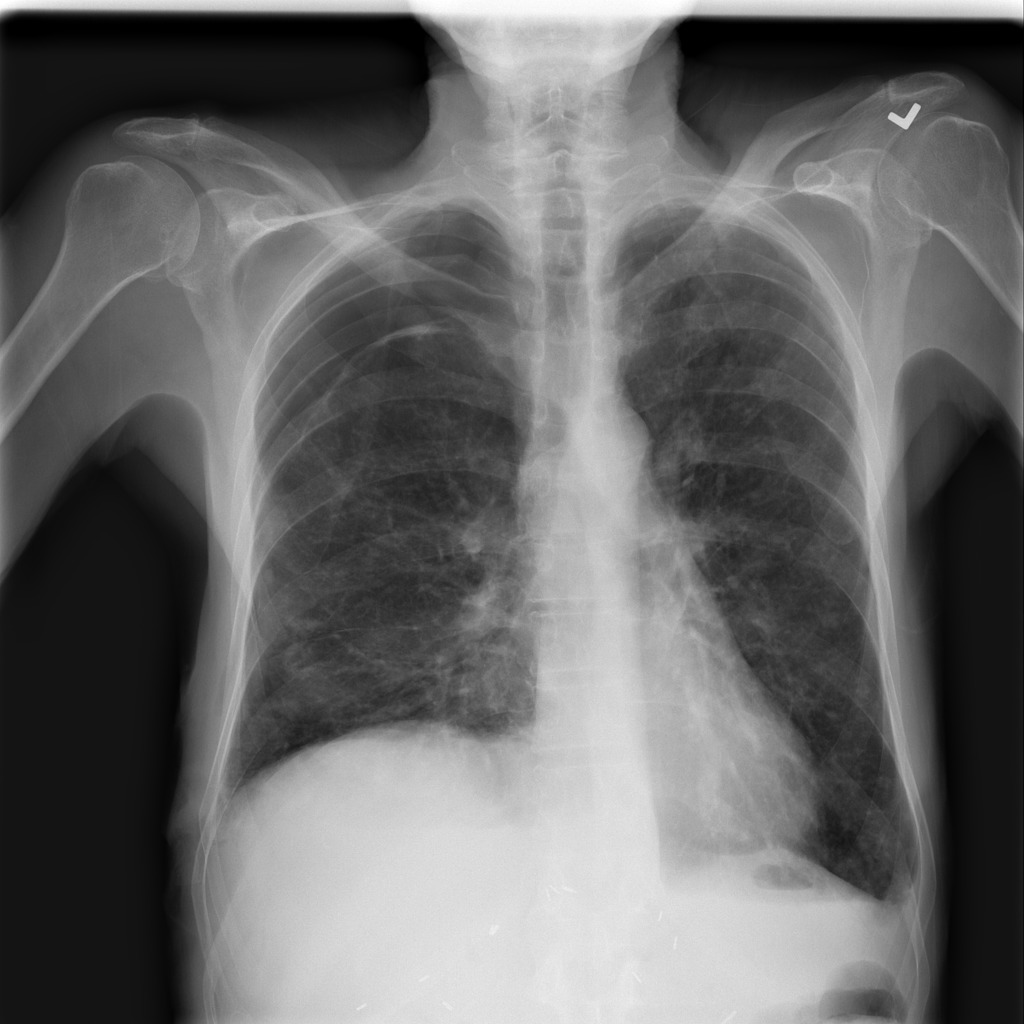}
    }
  \subfloat[Step1.CAM] {
    \includegraphics[width=0.18\linewidth]{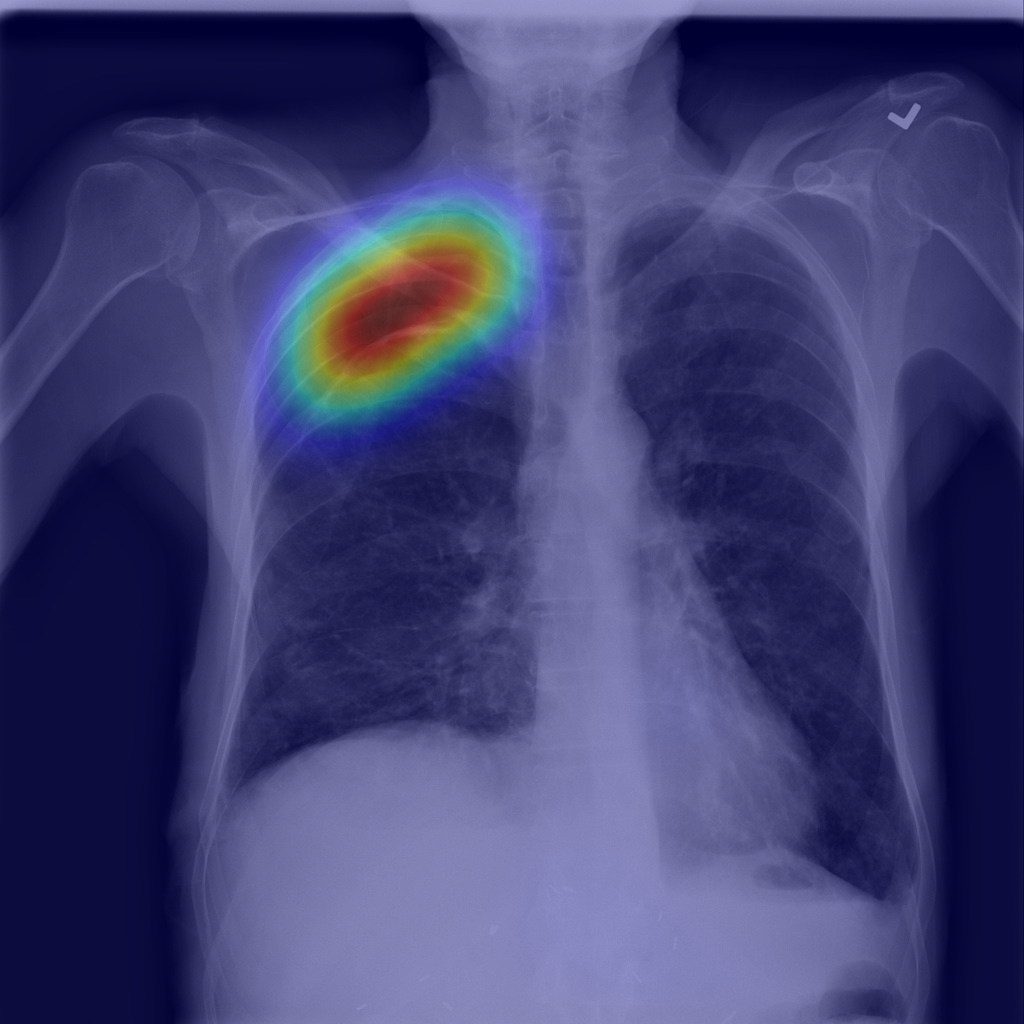}
    }
  \subfloat[Step2.IRNet] {
    \includegraphics[width=0.18\linewidth]{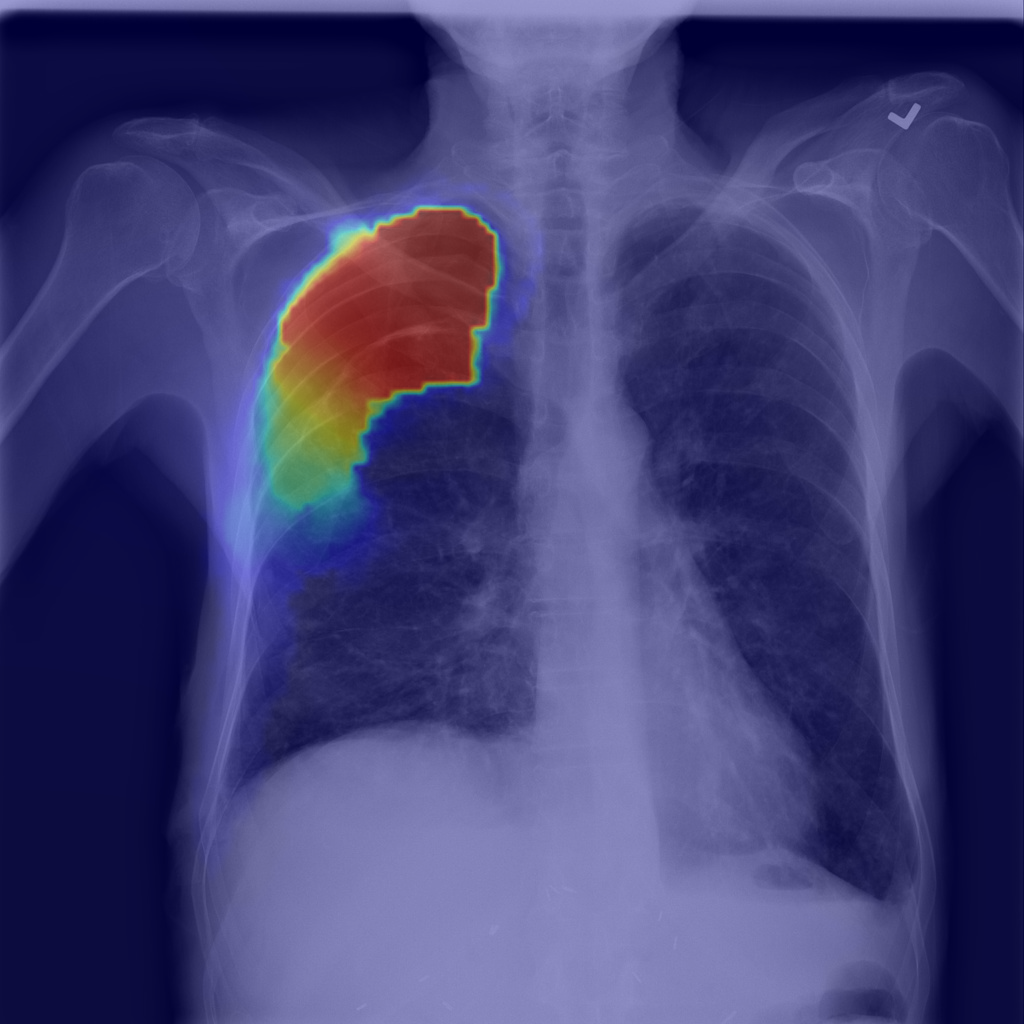}
    }
  \subfloat[Step3.Segm] {
    \includegraphics[width=0.18\linewidth]{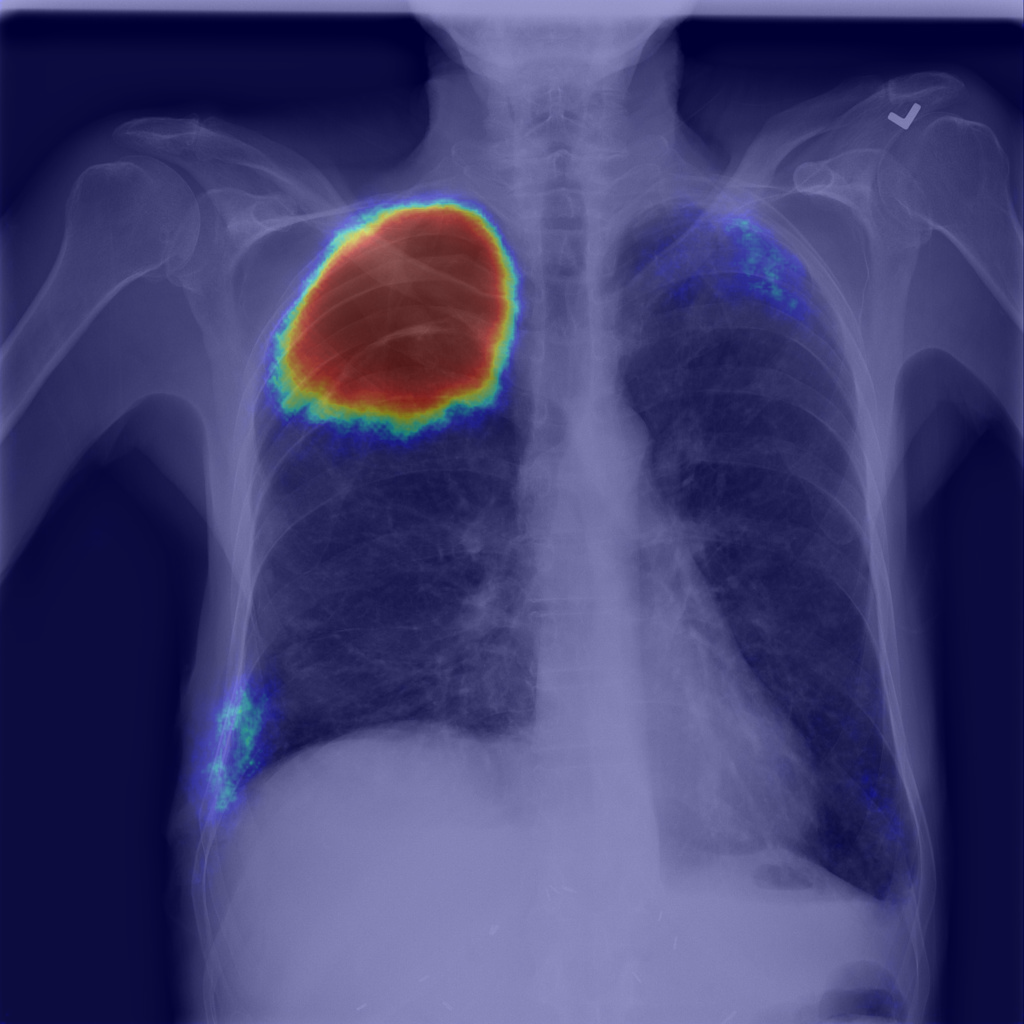}
    }
    \subfloat[Mask] {
    \includegraphics[width=0.18\linewidth]{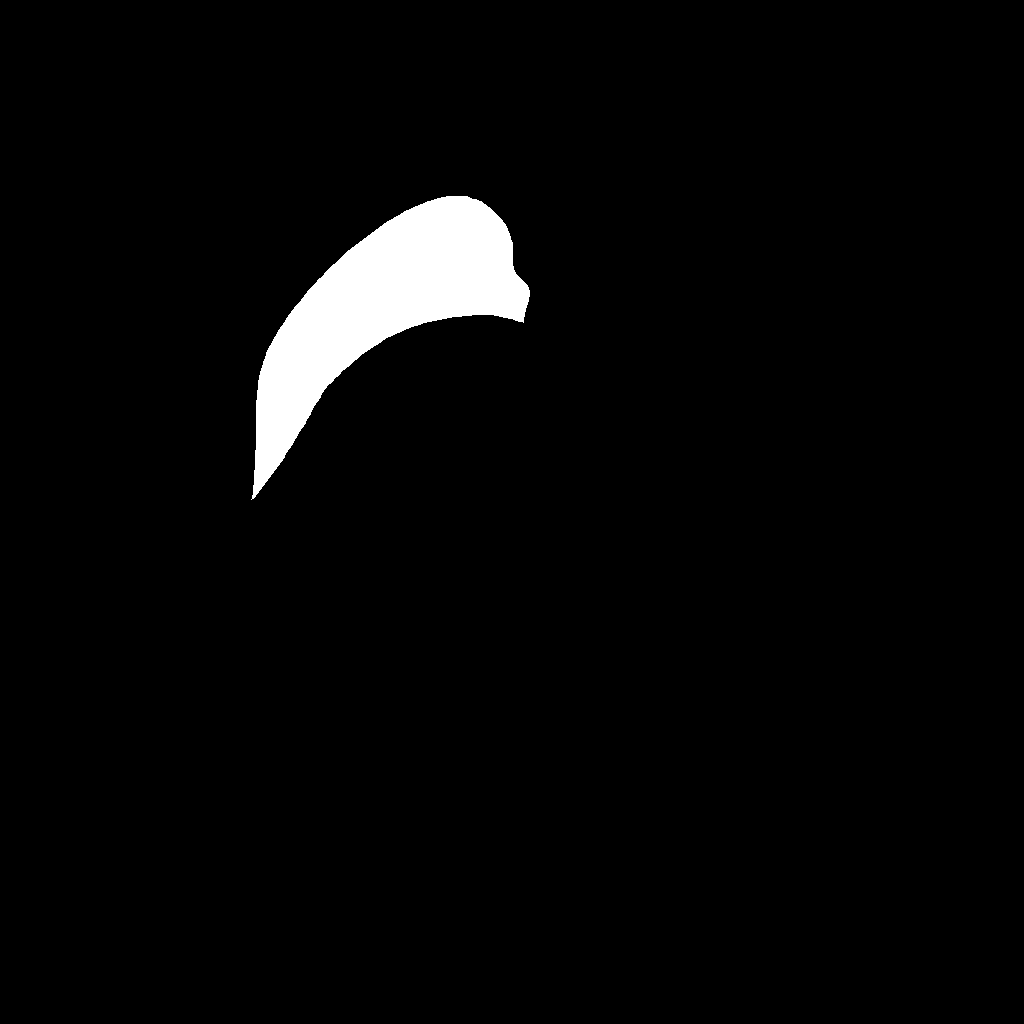}
    }
     \\[-1.5ex]
  \subfloat {
    \includegraphics[width=0.18\linewidth]{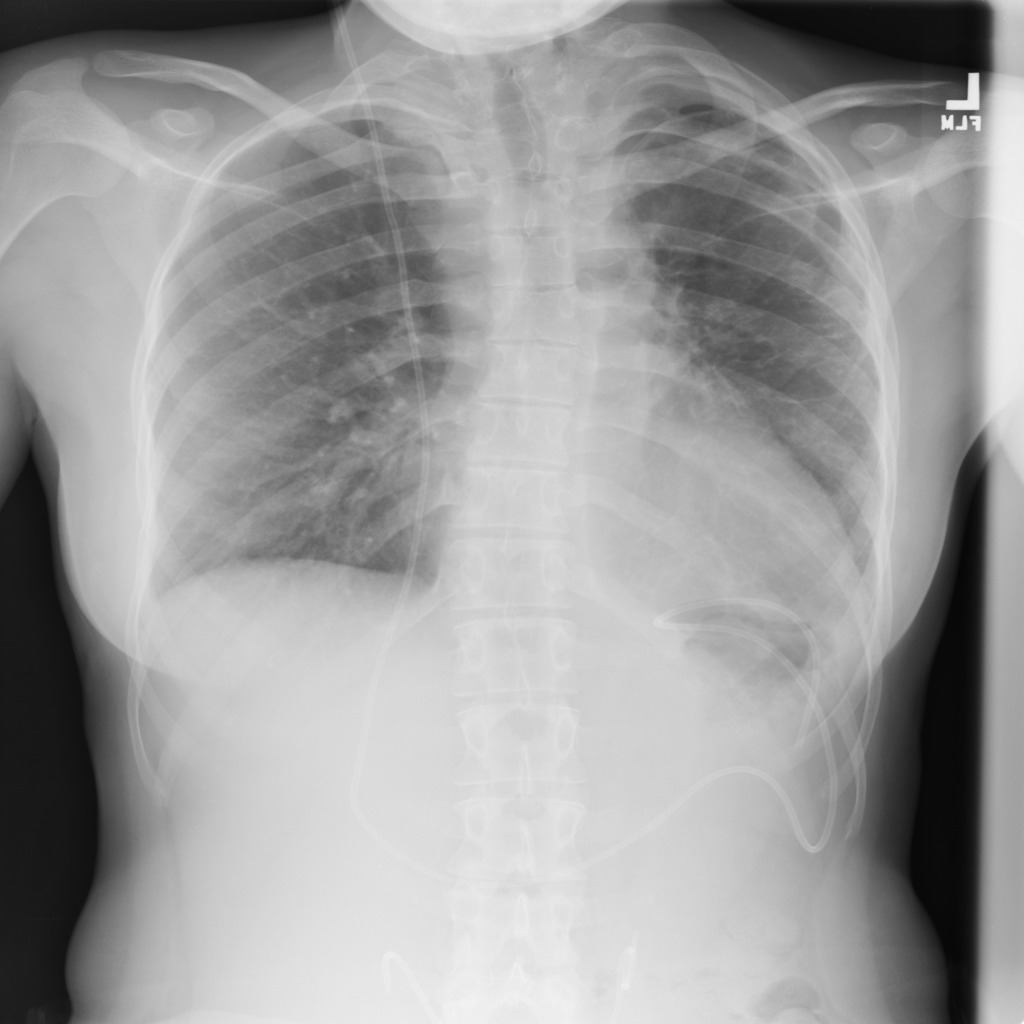}
    }
  \subfloat {
    \includegraphics[width=0.18\linewidth]{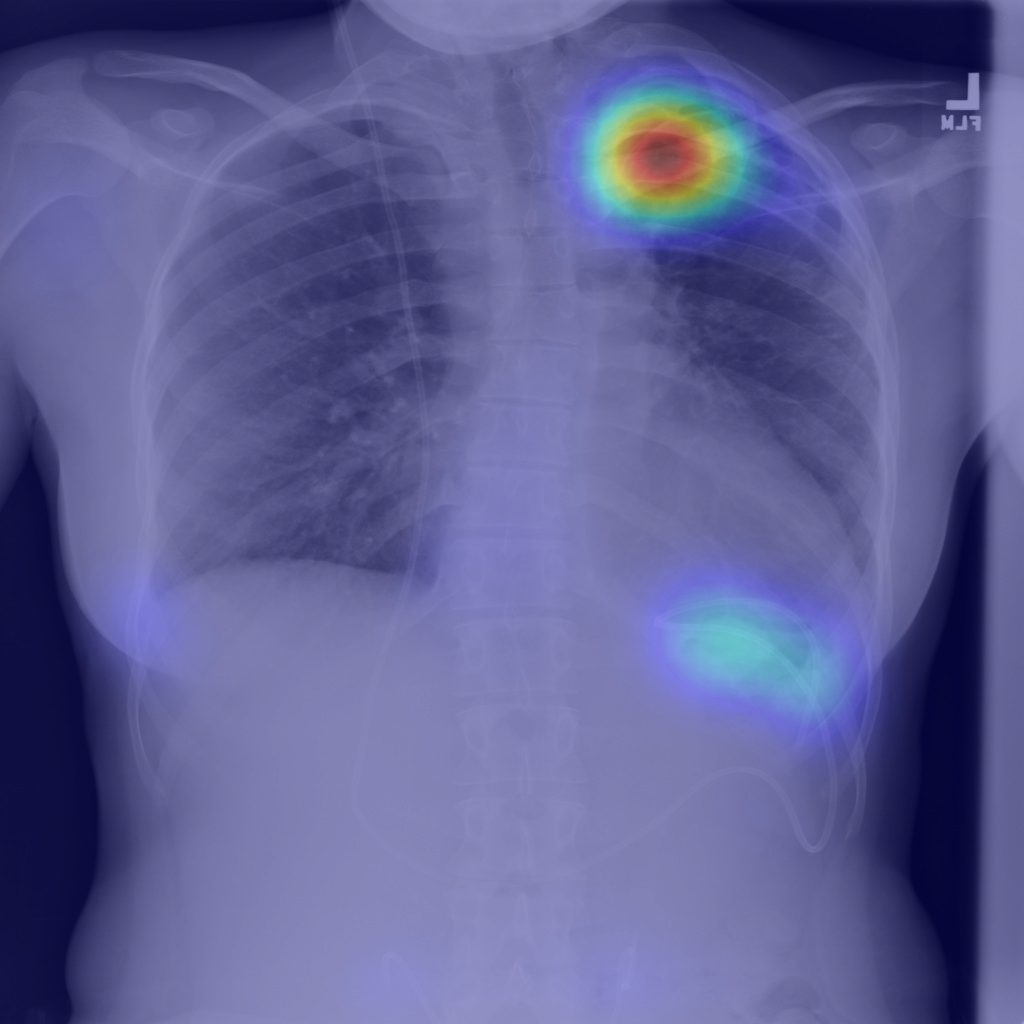}
    }
  \subfloat {
    \includegraphics[width=0.18\linewidth]{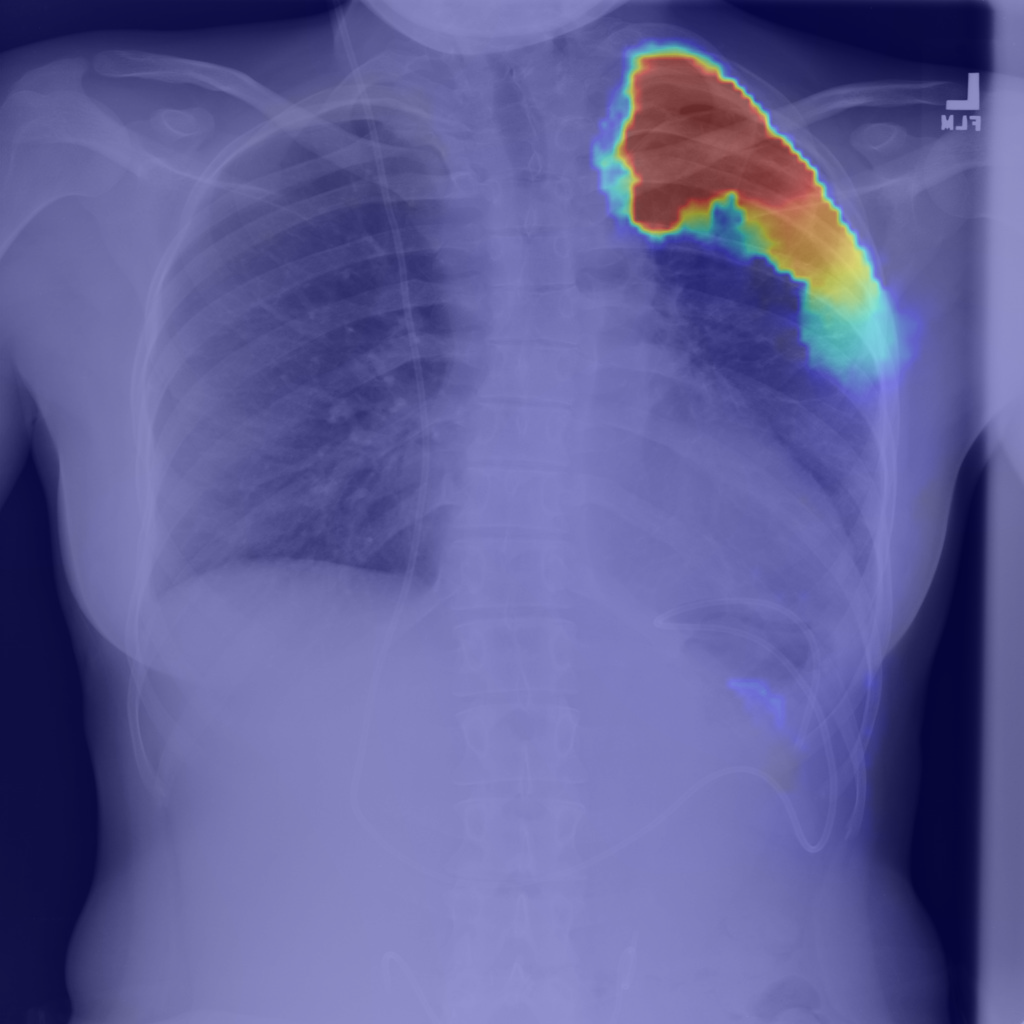}
    }
  \subfloat {
    \includegraphics[width=0.18\linewidth]{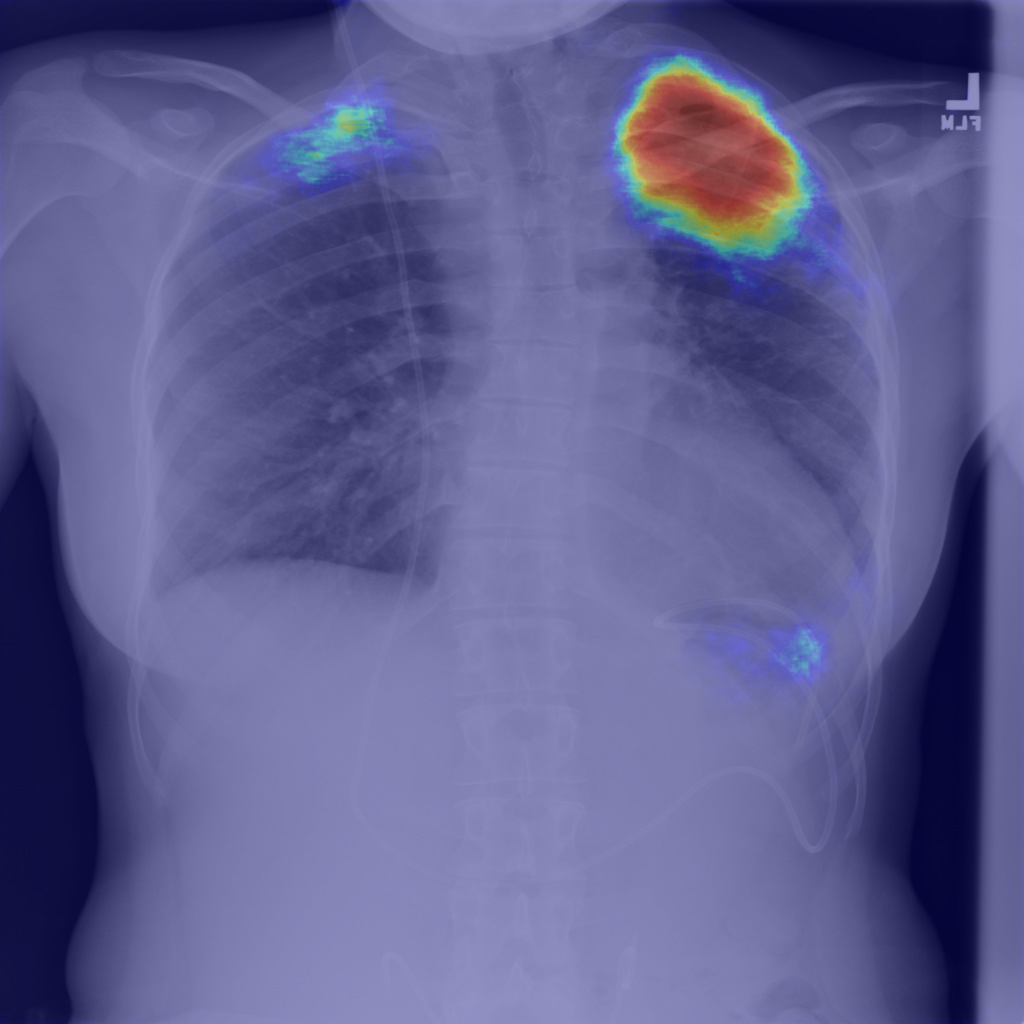}
    }
 \subfloat {
    \includegraphics[width=0.18\linewidth]{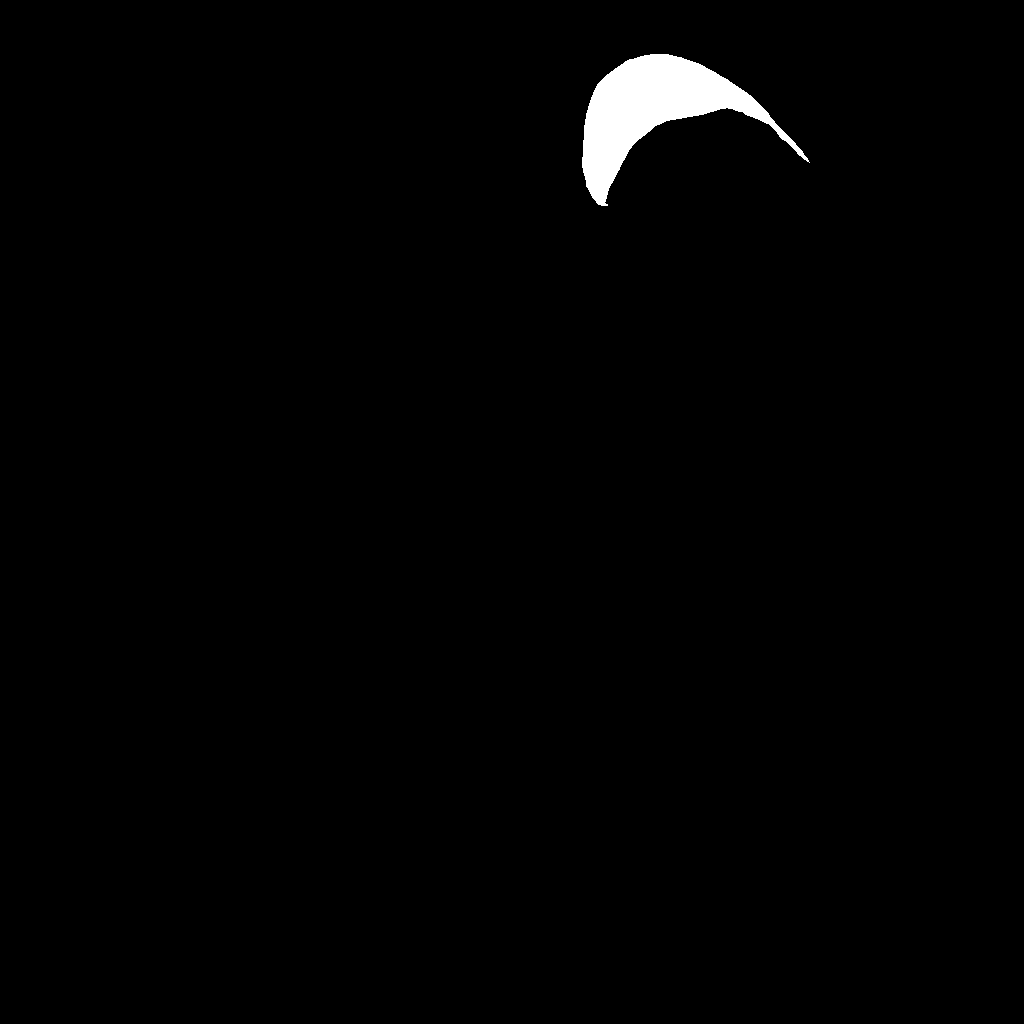}
    }
\caption{Pneumothorax localization maps for (a) a random image from the test set at each consecutive step of our method: (b) map after CAM extraction, (c) improved map by IRNet trained on the outcomed of step 1, (d) prediction of U-Net trained on step 2 results, all compared to (e) ground truth mask.}
\label{fig:siimlocalization}
\end{figure}

\begin{figure}[t!]
\vspace{-1.5em}
  \centering % <-- added
  \subfloat[Image] {
    \includegraphics[width=0.18\linewidth]{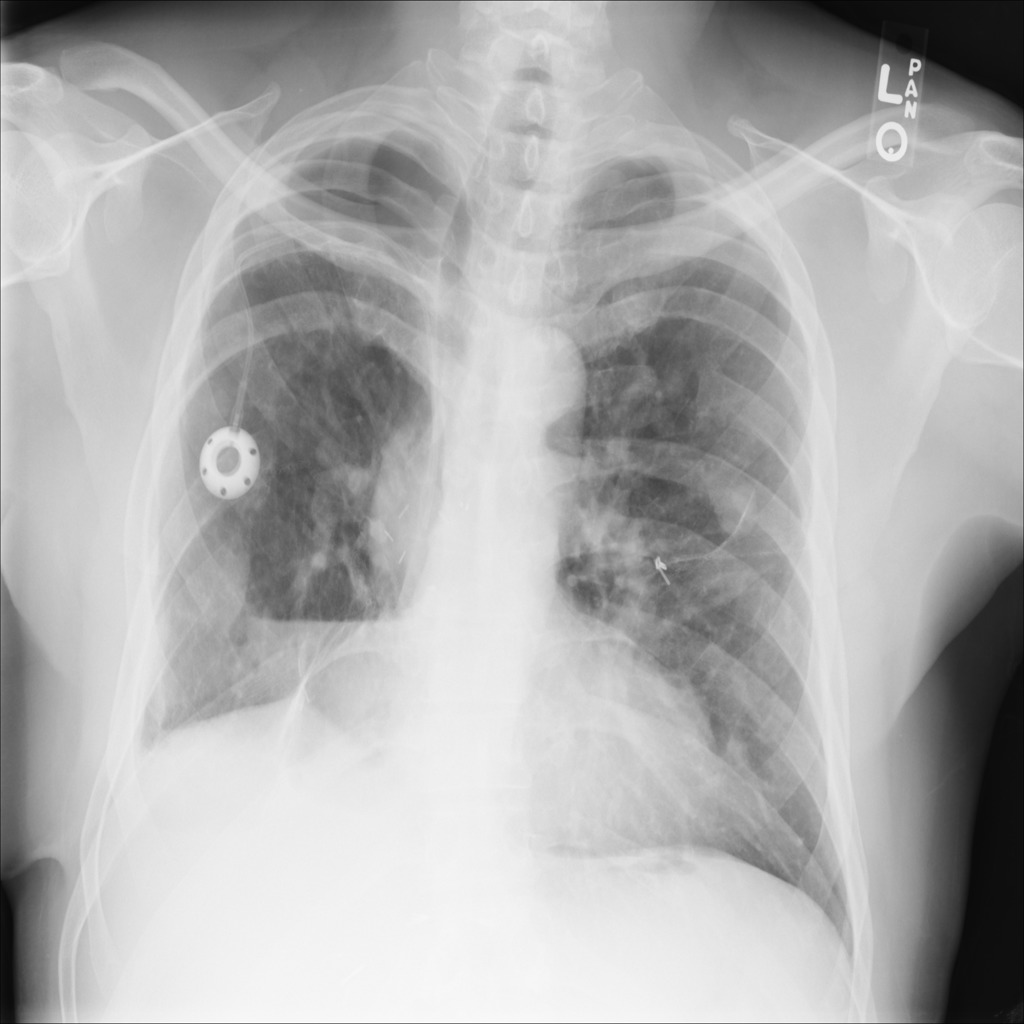}
    }
  \subfloat[Step1.CAM] {
    \includegraphics[width=0.18\linewidth]{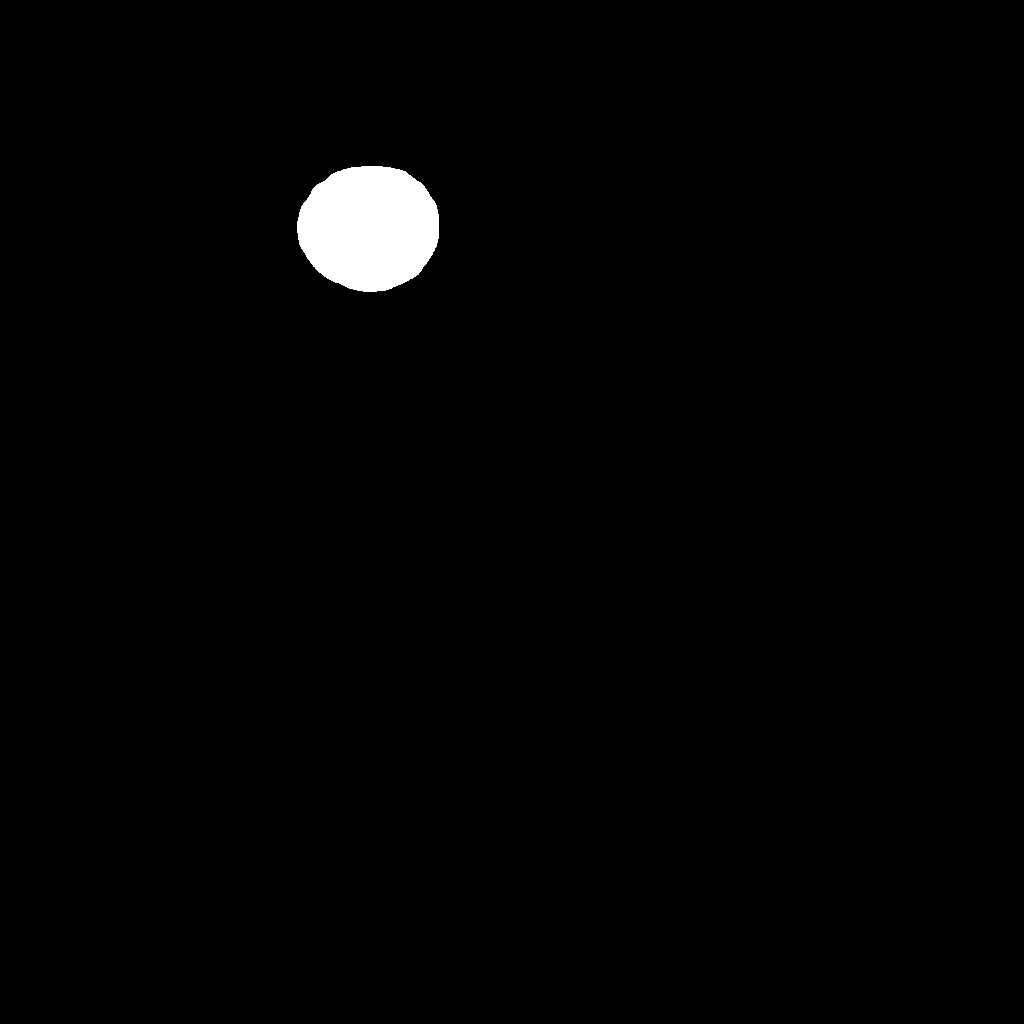}
    }
  \subfloat[Step2.IRNet] {
    \includegraphics[width=0.18\linewidth]{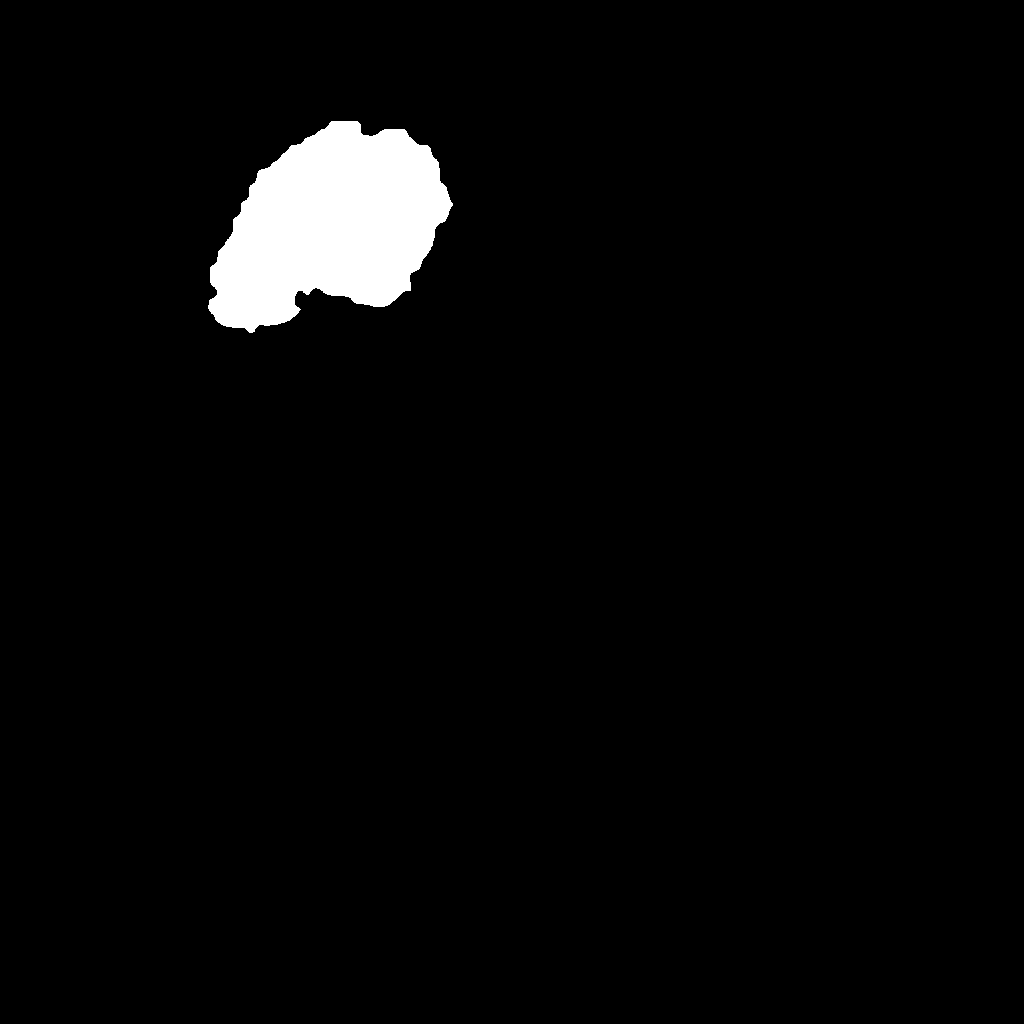}
    }
  \subfloat[Step3.Segm] {
    \includegraphics[width=0.18\linewidth]{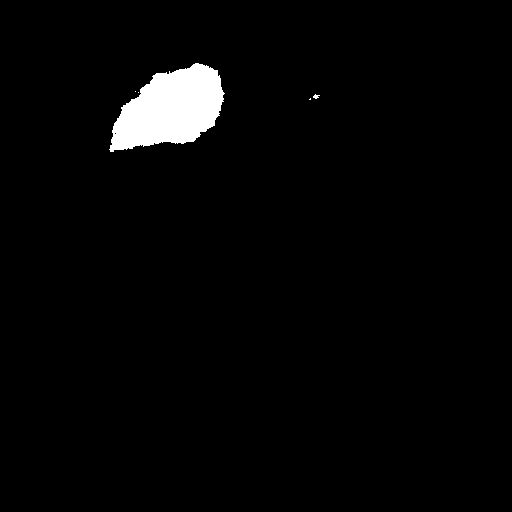}
    }
 \subfloat[Mask] {
    \includegraphics[width=0.18\linewidth]{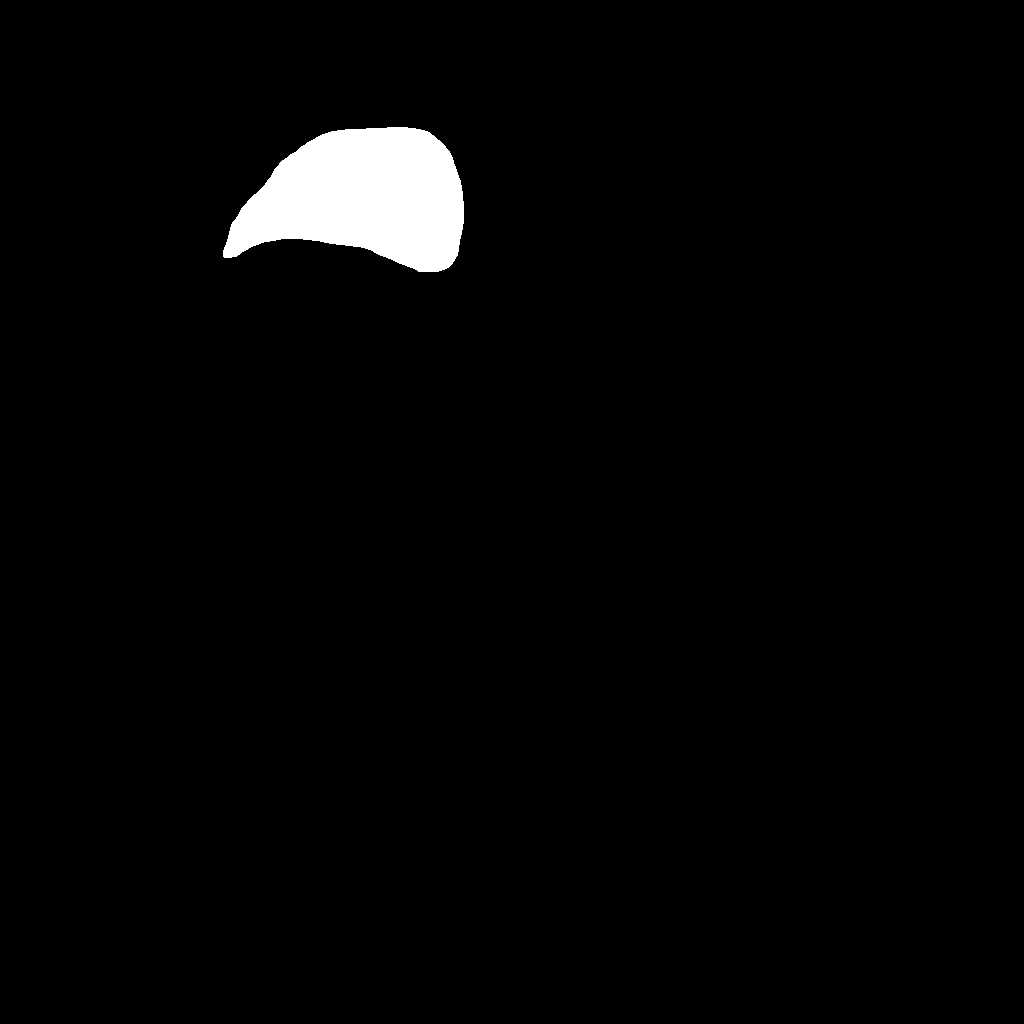}
    }
     \\[-1.5ex]
  \subfloat {
    \includegraphics[width=0.18\linewidth]{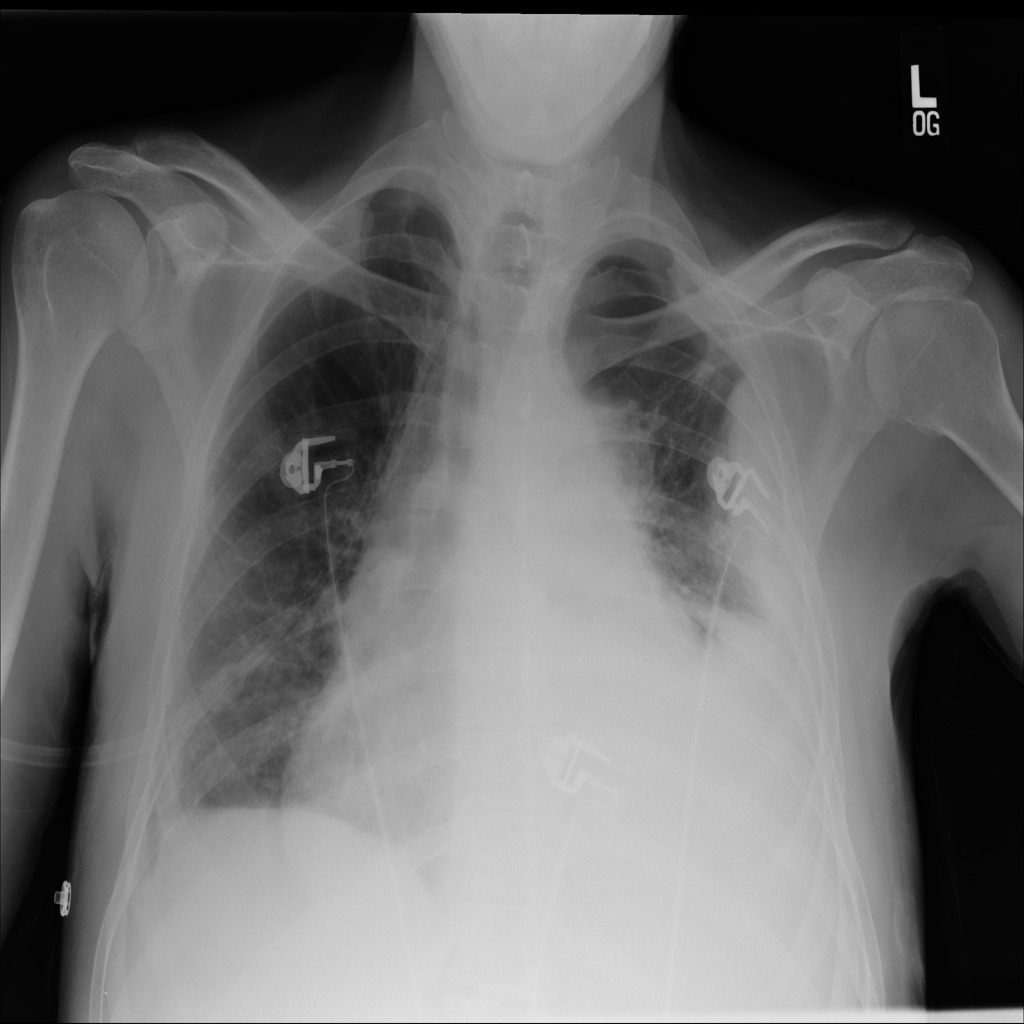}
    }
  \subfloat {
    \includegraphics[width=0.18\linewidth]{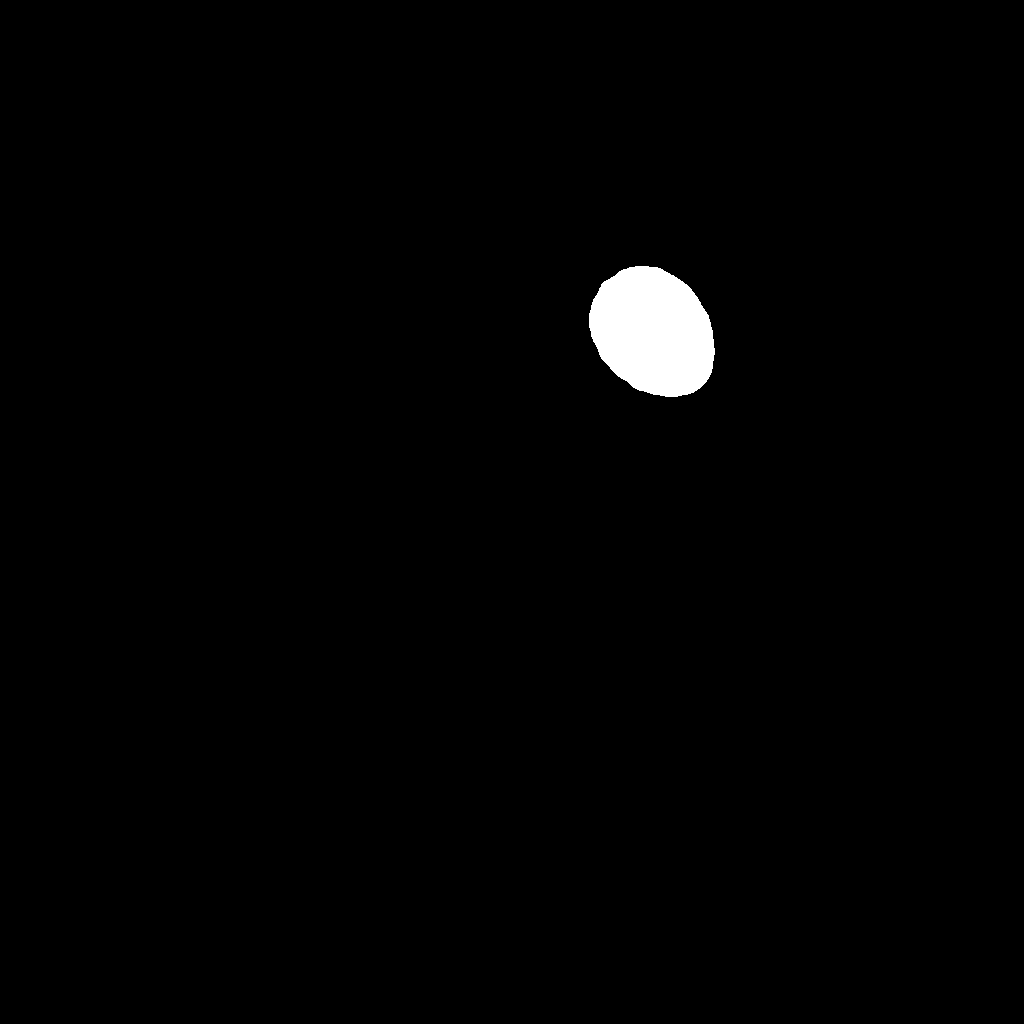}
    }
  \subfloat {
    \includegraphics[width=0.18\linewidth]{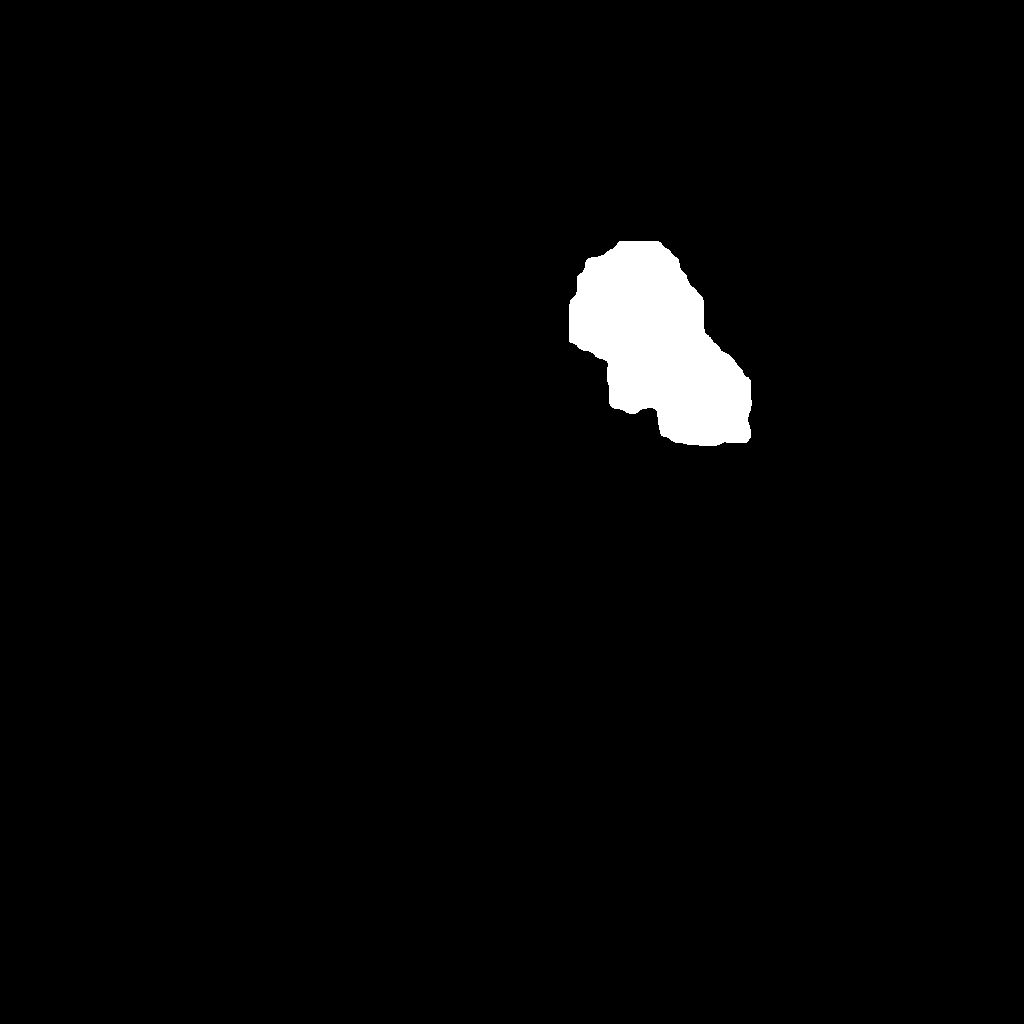}
    }
  \subfloat {
    \includegraphics[width=0.18\linewidth]{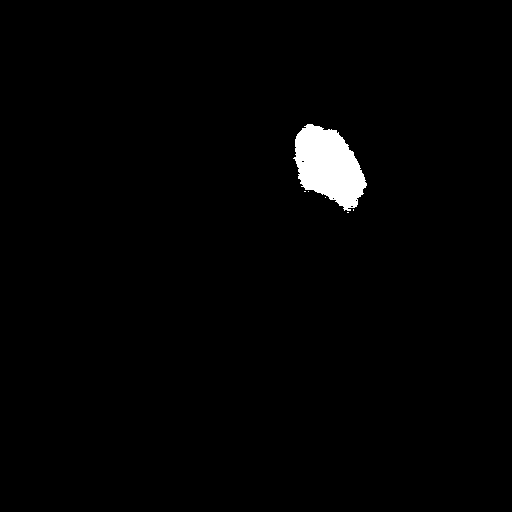}
    }
 \subfloat {
    \includegraphics[width=0.18\linewidth]{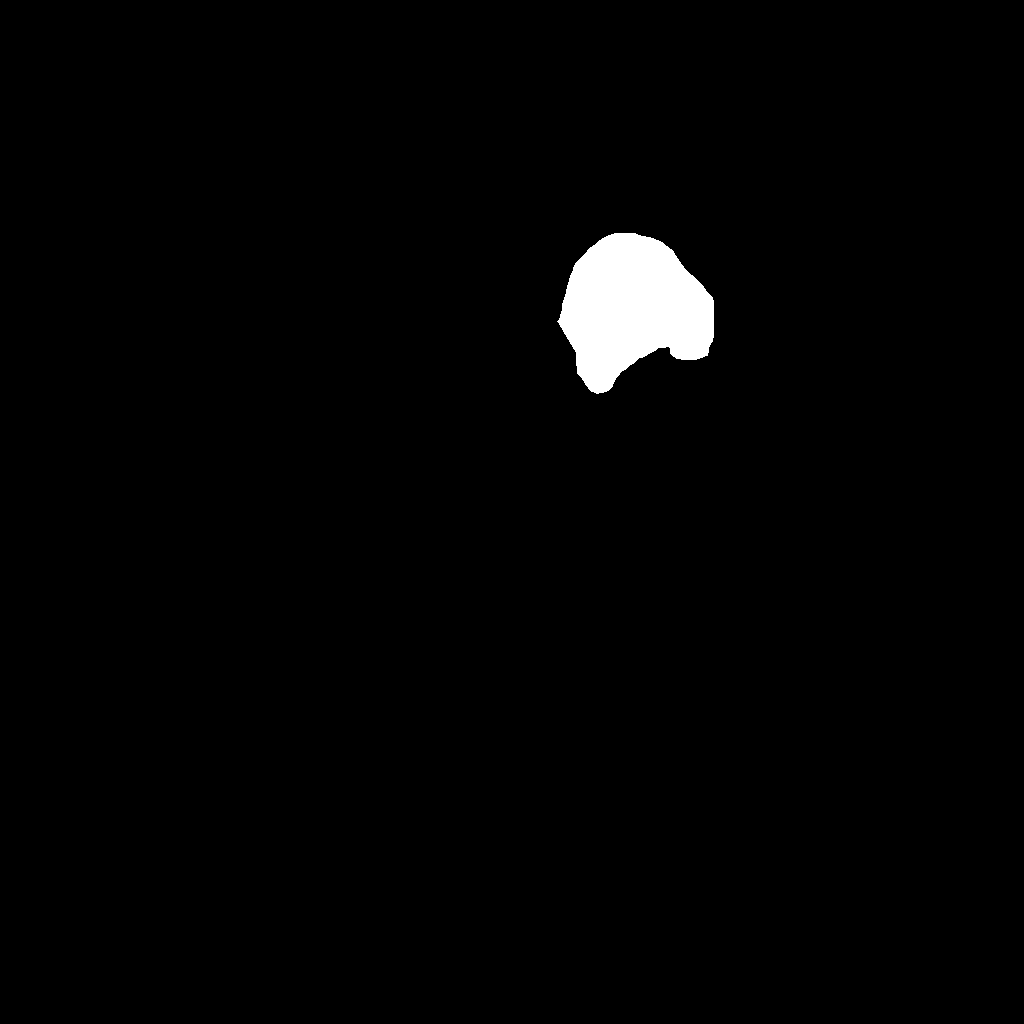}
    }
     \\[-1.5ex]
  \subfloat {
    \includegraphics[width=0.18\linewidth]{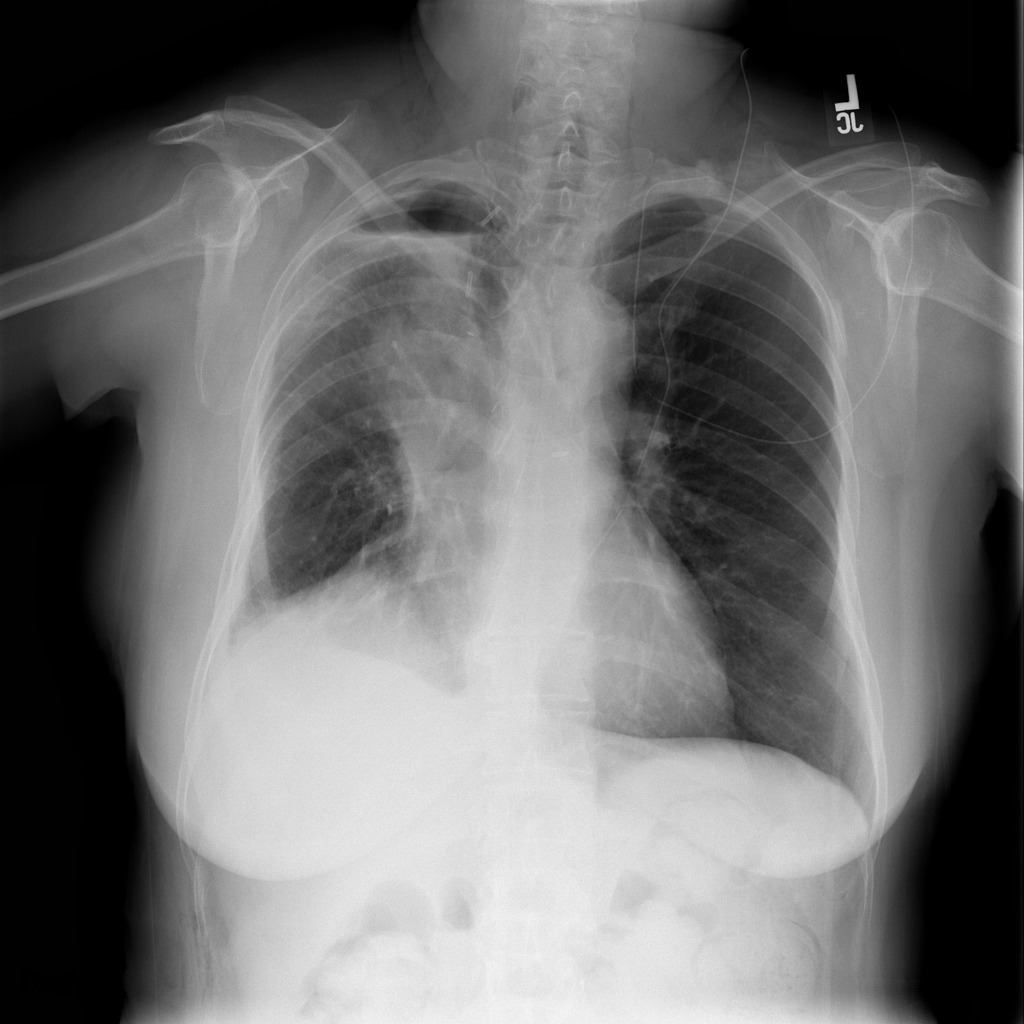}
    }
  \subfloat {
    \includegraphics[width=0.18\linewidth]{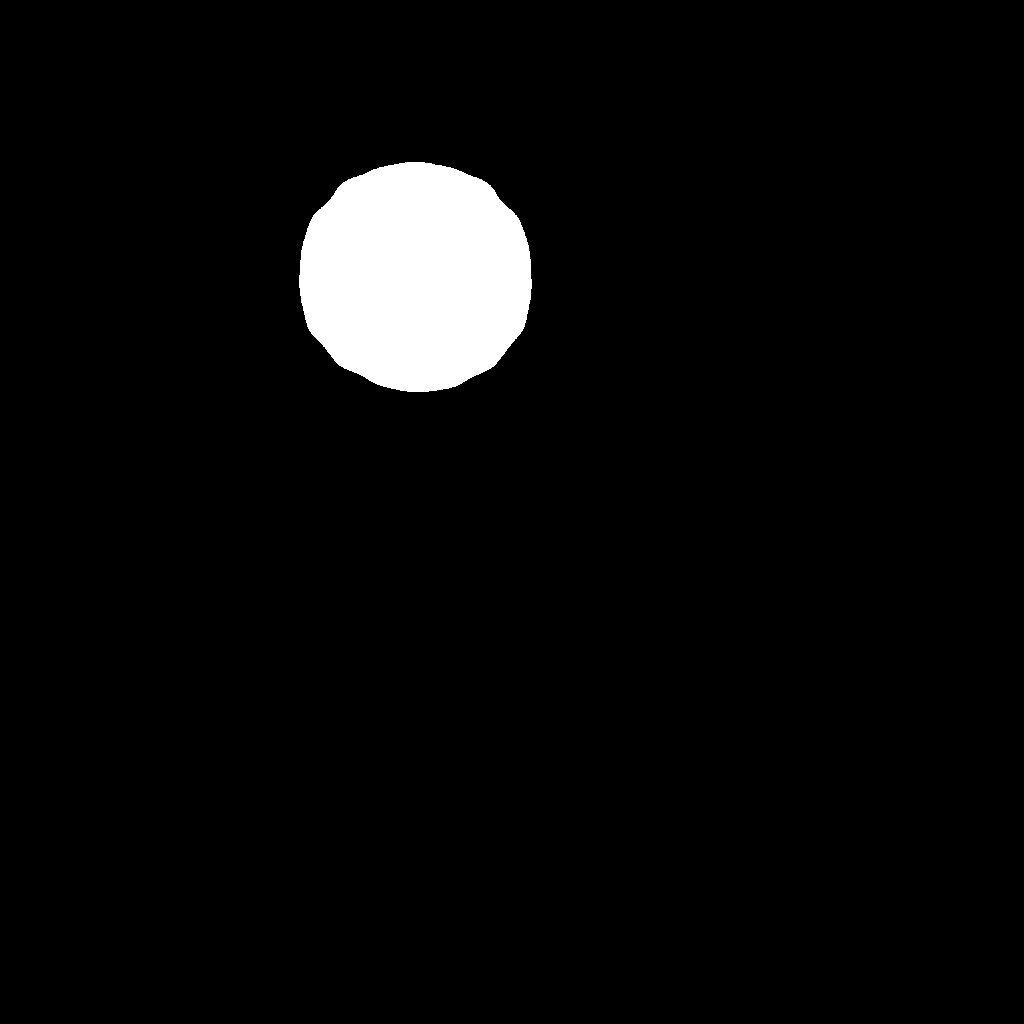}
    }
  \subfloat {
    \includegraphics[width=0.18\linewidth]{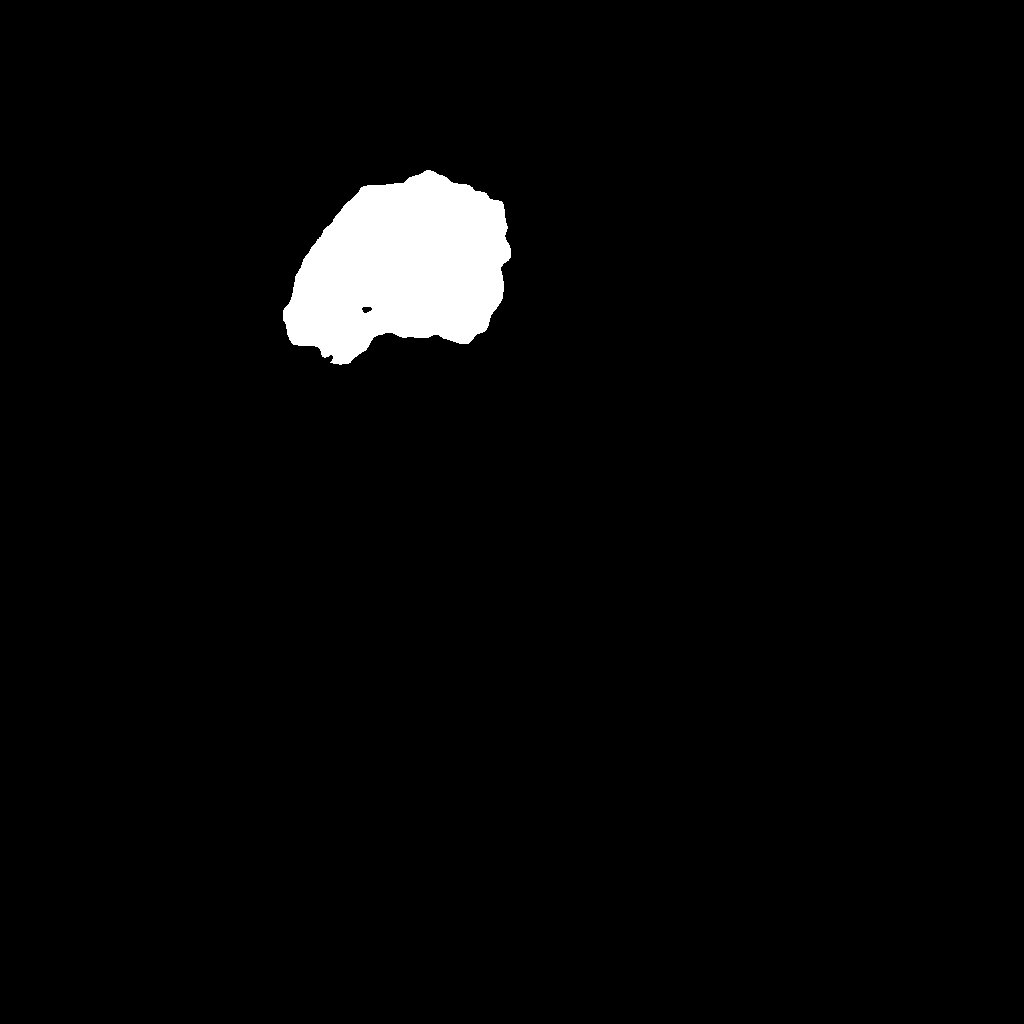}
    }
  \subfloat {
    \includegraphics[width=0.18\linewidth]{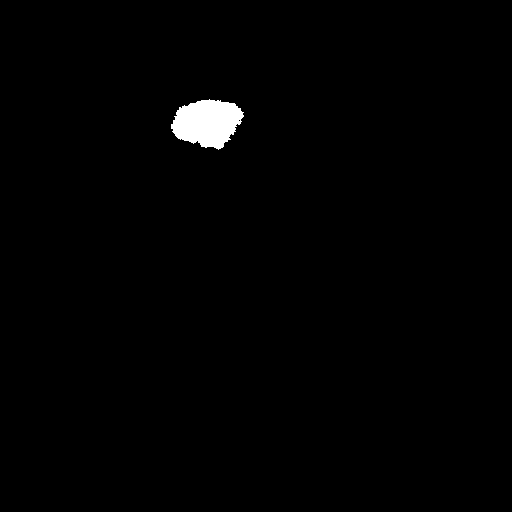}
    }
 \subfloat {
    \includegraphics[width=0.18\linewidth]{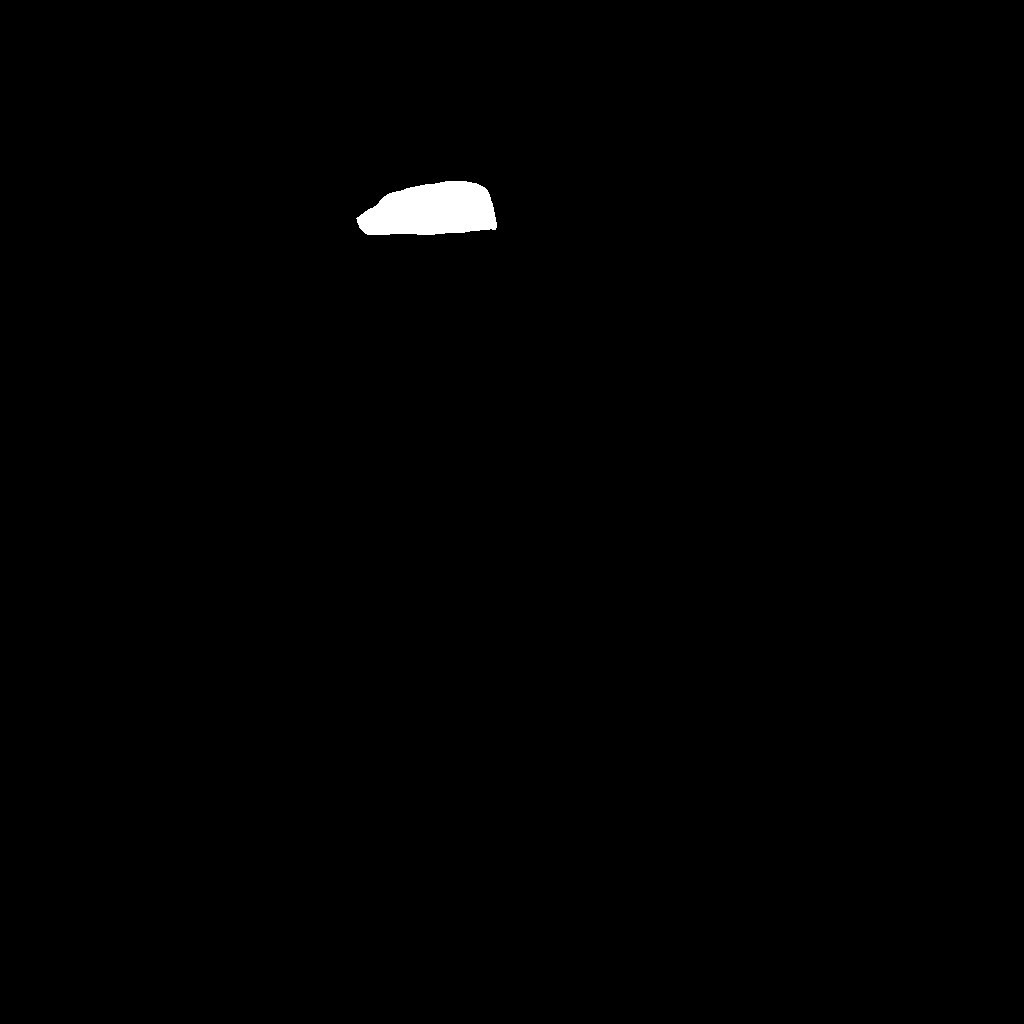}
    }
%  ----------------------------------------------------------------
\caption{Segmentation predictions for (a) a random image from test set of SIIM-ACR Pneumothorax produced at each step of our approach: (b) CAM extraction, (c) IRNet, (d) U-Net segmentation, compared to (e) ground truth mask.}
\label{fig:siimimages}
% \vspace{-5em}
\end{figure}

\begin{figure}[H]
  \centering % <-- added
  \subfloat[Image] {
    \includegraphics[width=0.18\linewidth]{images/pascal/2008_002835_image.jpg}
    }
  \subfloat[Step1.CAM] {
    \includegraphics[width=0.18\linewidth]{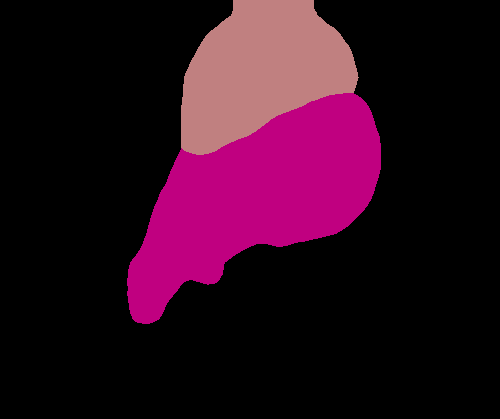}
    }
  \subfloat[Step2.IRNet] {
    \includegraphics[width=0.18\linewidth]{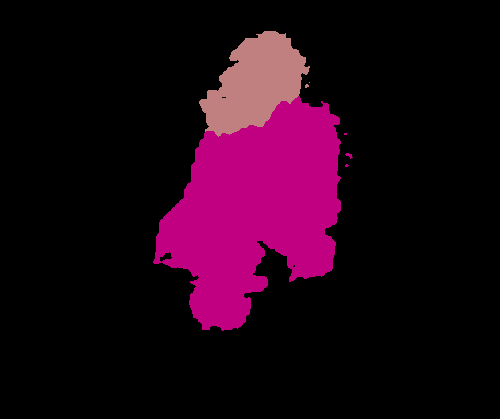}
    }
  \subfloat[Step3.Segm] {
    \includegraphics[width=0.18\linewidth]{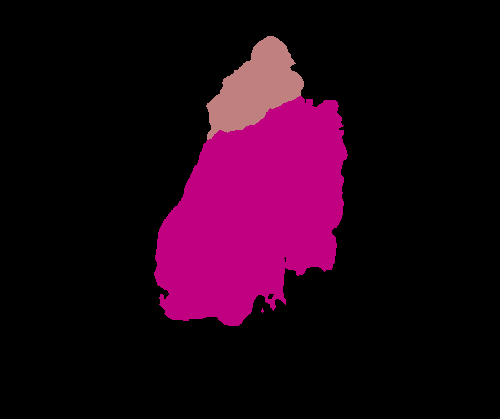}
    }
    \subfloat[Mask] {
    \includegraphics[width=0.18\linewidth]{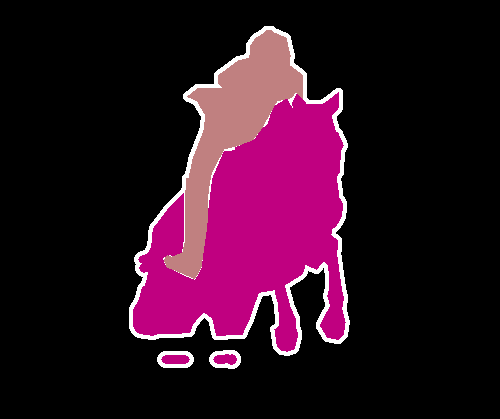}
    }
    \\[-1.5ex]
  \subfloat {
    \includegraphics[width=0.18\linewidth]{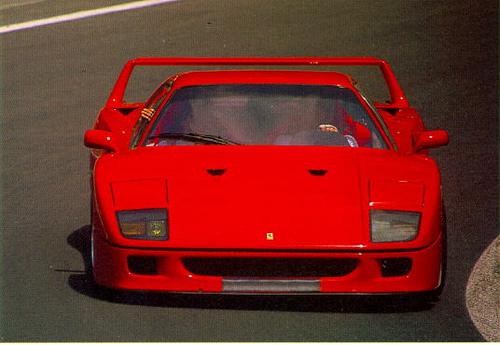}
    }
  \subfloat {
    \includegraphics[width=0.18\linewidth]{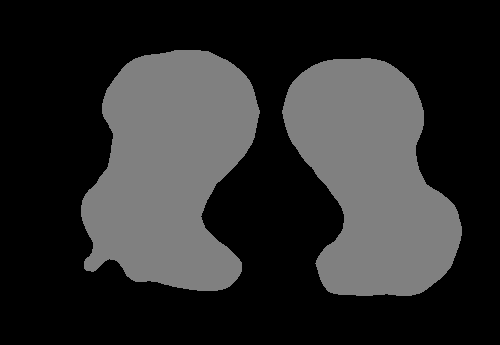}
    }
  \subfloat {
    \includegraphics[width=0.18\linewidth]{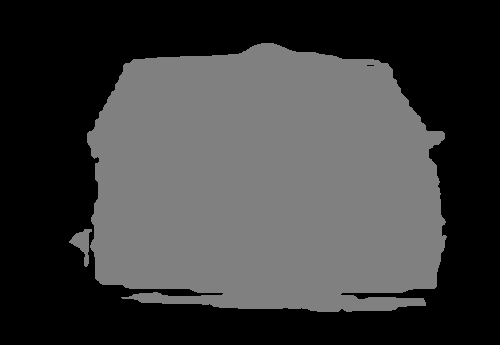}
    }
  \subfloat {
    \includegraphics[width=0.18\linewidth]{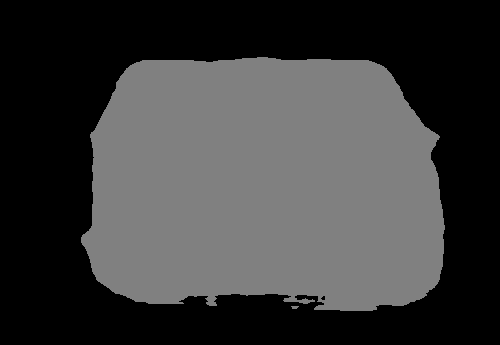}
    }
 \subfloat {
    \includegraphics[width=0.18\linewidth]{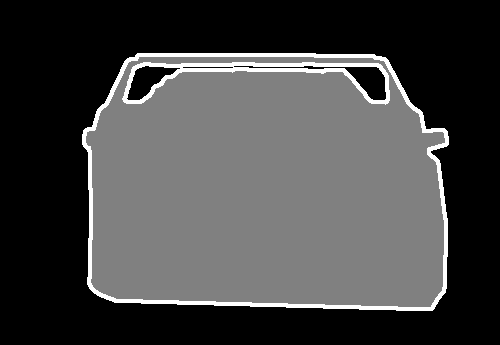}
    }
     \\[-1.5ex]
  \subfloat {
    \includegraphics[width=0.18\linewidth]{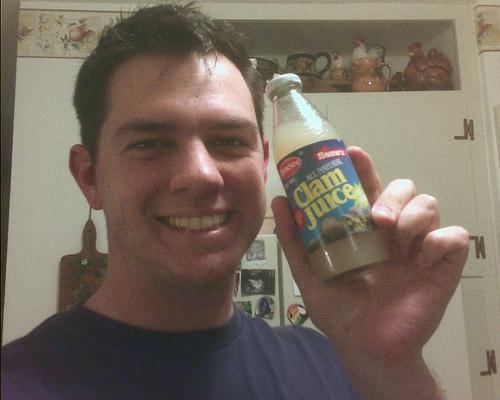}
    }
  \subfloat {
    \includegraphics[width=0.18\linewidth]{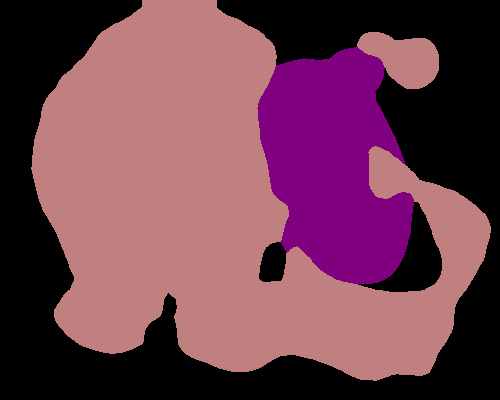}
    }
  \subfloat {
    \includegraphics[width=0.18\linewidth]{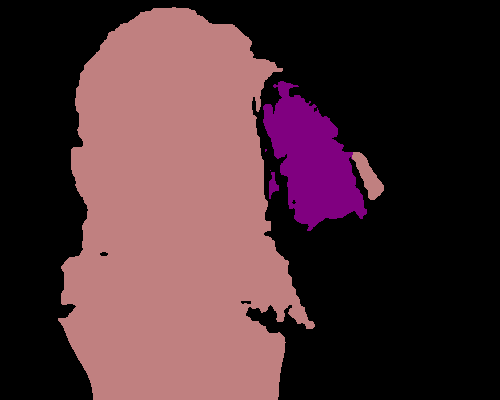}
    }
  \subfloat {
    \includegraphics[width=0.18\linewidth]{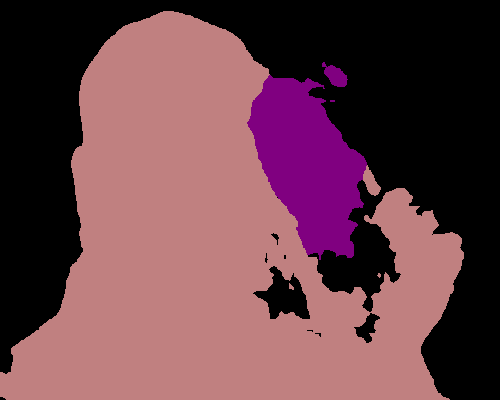}
    }
 \subfloat {
    \includegraphics[width=0.18\linewidth]{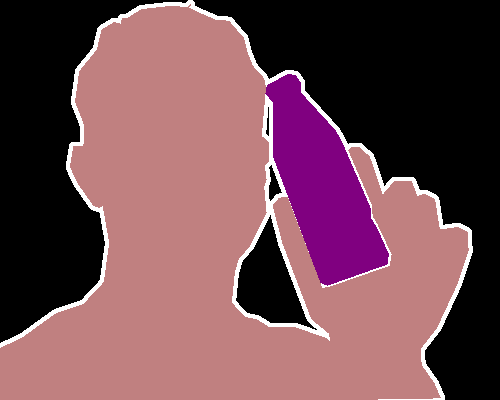}
    }
    \\[-1.5ex]
  \subfloat {
    \includegraphics[width=0.18\linewidth]{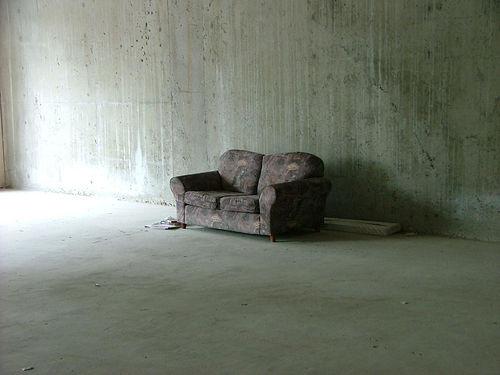}
    }
  \subfloat {
    \includegraphics[width=0.18\linewidth]{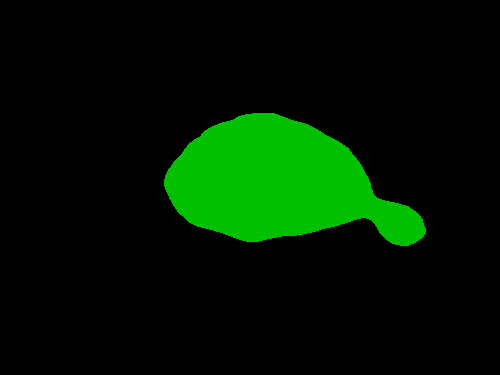}
    }
  \subfloat {
    \includegraphics[width=0.18\linewidth]{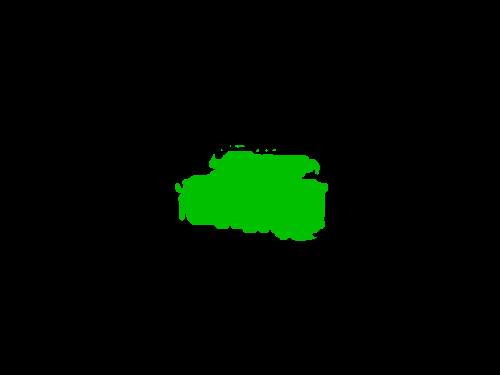}
    }
  \subfloat {
    \includegraphics[width=0.18\linewidth]{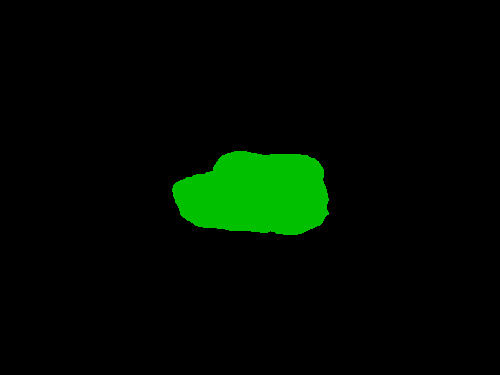}
    }
 \subfloat {
    \includegraphics[width=0.18\linewidth]{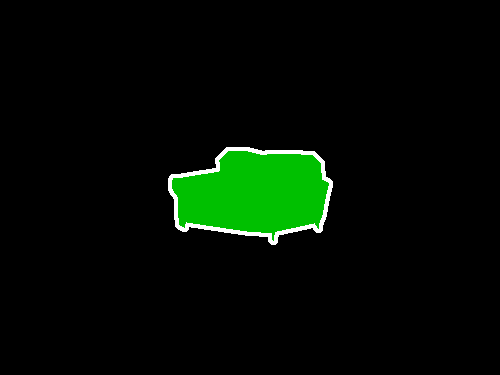}
    }
    \\[-1.5ex]
  \subfloat {
    \includegraphics[width=0.18\linewidth]{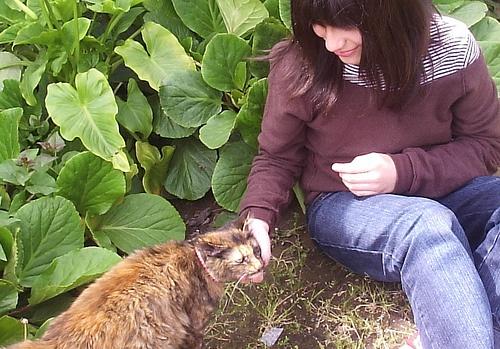}
    }
  \subfloat {
    \includegraphics[width=0.18\linewidth]{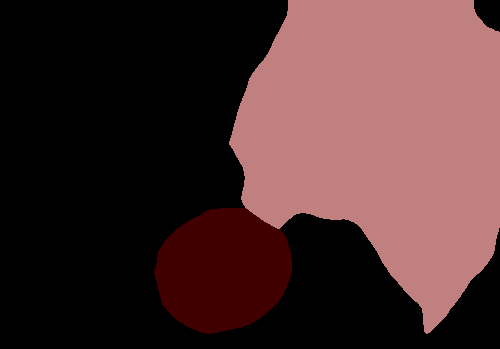}
    }
  \subfloat {
    \includegraphics[width=0.18\linewidth]{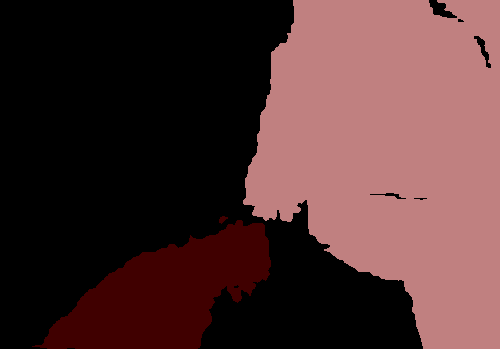}
    }
  \subfloat {
    \includegraphics[width=0.18\linewidth]{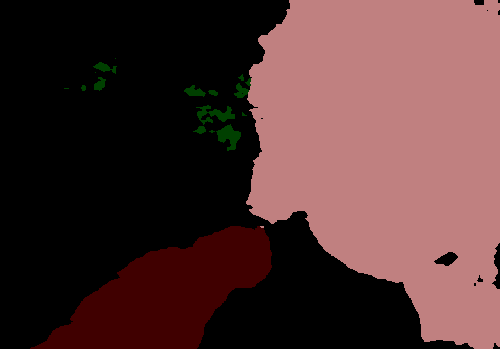}
    }
 \subfloat {
    \includegraphics[width=0.18\linewidth]{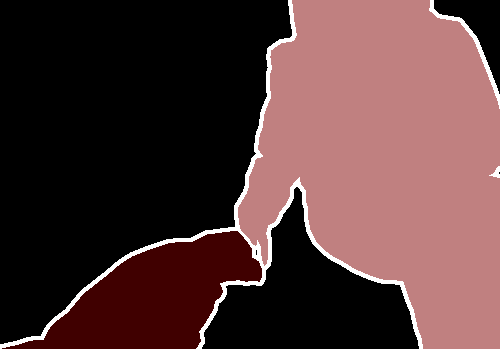}
    }

\caption{Visualization of segmentation predictions on PASCAL VOC 2012 for (a) an image produced at each step of our approach: (b) CAM extraction, (c) IRNet, (d) DeepLabv3+ segmentation, compared to (e) ground truth mask.}
\label{fig:pascalevl}
\end{figure}

We present method's explainability via disease localization regions; cf. Figure \ref{fig:siimlocalization}. We provide qualitative results of segmentation on validation images from both da\-ta\-sets in Figure \ref{fig:siimimages} and Figure \ref{fig:pascalevl}. We show the resulting maps at each step of our method; the figures demonstrate how the performance improves after each step. We achieve comparable results to state-of-the-art method on PASCAL VOC 2012; cf. Table \ref{pascalresults}. 
\par
We evaluate our method on the second stage test set on the competition server \cite{siimpneumothorax} to compare it against a fully-supervised upper-performance limit. We achieve 0.769 Dice score while the first place solution got 0.868 using pixel-level labels for training. Our method proves the capability of using only image-level annotations for semantic segmentation on chest X-rays, nevertheless, attaining as good or even better results than those produced by fully supervised networks is still a challenge for weakly-supervised approaches.

\vspace{-2em}
% \begin{table}
% \centering
% \caption{Comparison of weakly-supervised semantic segmentation methods on PASCAL VOC 2012 validation set. Our approach is evaluated after each of the proposed steps, where each step is trained on the outcomes of the previous one. The final results are reported as Ours. Step3. Segmentation.}
% \vspace{1.0em}
% \begin{tabularx}{\textwidth}{l|P{1cm}|l|P{1.5cm}}
% \hline
% Method & Year & Model &  mIoU \\
% % \hline
% \specialrule{.1em}{.05em}{.05em} 
% Ours. Step1. CAM & 2020 & VGG16 and Grad-CAM++  & 0.479 \\ \hline
% Ours. Step2. IRNet & 2020 & ResNet50 after CAM generation & 0.631 \\ \hline

% \multirow{}{3cm}{Ours. Step3. Segmentation } & 2020 & DeepLabv3+ - ResNet50 after          CAM & 0.505 \\ 
%         & 2020 &  DeepLabv3+ - ResNet50 after IRNet & \bfseries 0.646 \\

% \specialrule{.1em}{.05em}{.05em} 
% IRNet \cite{ahn2019weakly} & 2019 & DeepLab-v2-ResNet50  & 0.635 \\ \hline
% FickleNet \cite{lee2019ficklenet} & 2019 & DeepLab-v2-ResNet101 and dropout - 0.9  & \bfseries 0.649 \\ \hline
% DSRG (ResNet101) \cite{huang2018weakly} & 2018 & DeepLab-ASPP & 0.614 \\ \hline
% SEC \cite{kolesnikov2016seed} & 2016 & DeepLab-CRF-LargeFOV  & 0.507  \\ 
% \hline
% \end{tabularx}
% \label{pascalresults}
% \end{table}

\begin{table}
\centering
\caption{Comparison of weakly-supervised semantic segmentation methods on PASCAL VOC 2012 validation set. Our approach is evaluated after each of the proposed steps, where each step is trained on the outcomes of the previous one.}
\vspace{1.0em}
\begin{tabularx}{\textwidth}{l|P{3.5cm}|P{3.5cm}}
\hline
Method & Year &  mIoU \\
% \hline
\specialrule{.1em}{.05em}{.05em} 
Our method.  Step 1. CAM & 2020  & 0.479 \\ \hline
Our method.  Step 2. IRNet & 2020 & 0.631 \\ \hline
Our method.  Step 3. Segmentation & 2020 & \bfseries 0.646  \\ 
\specialrule{.1em}{.05em}{.05em} 
IRNet \cite{ahn2019weakly} & 2019 & 0.635 \\ \hline
FickleNet \cite{lee2019ficklenet} & 2019 & \bfseries 0.649 \\ \hline
DSRG (ResNet101) \cite{huang2018weakly} & 2018 & 0.614 \\ \hline
SEC \cite{kolesnikov2016seed} & 2016  & 0.507  \\ 
\hline
\end{tabularx}
\label{pascalresults}
\end{table}

\vspace{-4em}

\section{Clinical Relevance}

During diagnosing procedure the final decision maker is a doctor, while AI-powered decision support systems can assist by detecting regions of interest and presenting the data in a convenient format that doctors can use. With an automatic image segmentation solution, the healthcare provider can reach higher efficiency by saving doctors' time spent on the primary analysis of images. At the same time it will increase diagnosis accuracy by providing the second opinion. The major problem in building such a solution is the lack of large amounts of pixel-wise labeled data that are extremely costly in terms of the expert time required for their annotation. With our approach, which requires only image-level annotations, the costs can be reduced dramatically. In the long run, it allows the cheaper implementation of segmentation models, also facilitating research in the area by overcoming the problem of collecting datasets with pixel-wise annotations. Saving doctors' time on diagnosis is especially important during the disease widespread such as COVID-19, when large number of people are affected by a disease and the amount of required screening procedures grows exponentially. 
\par
Using our method for medical images can automate parts of the radiology workflow cutting operational costs for the hospitals. The proposed approach was designed to be general and applicable to other medical purposes; for example detection of various thoracic diseases. 

\section{Conclusions}
% We present a novel method for weakly-supervised semantic segmentation, which is effective not only for natural domain images but to medical as well. The presented approach consists of two steps responsible for the pseudo-label generation and followed by a supervised segmentation model. This allows training a segmentation model with limited supervision. Only image-level annotations are used to predict the location of Pneumothorax on chest X-ray images. The performance on SIIM-ACR Pneumothorax proves the method's ability to produce accurate pixel-level disease localization maps. Our approach also shows comparable results to state-of-the-art methods on PASCAL VOC 2012. The developed approach saves resources required for pixel-wise labeling by enabling the use of image-level annotations for segmentation task. 

We present a novel method of weakly-supervised semantic segmentation that demonstrated its efficiency for detecting anomalous regions on chest X-ray images. In particular, we propose a three-step approach to weakly-supervised semantic segmentation, which uses only image-level labels as supervision. Next, we customize and expand the previous works by including supplementary steps such as regularization, IRNet, and various post-processing techniques. Also, the method is general, domain independent and explainable via localization maps at each step. We evaluated it on two datasets of different nature; however, it can also be implemented in other medical problems.

\section*{Acknowledgements}

This research was supported by SoftServe and Faculty of Applied Sciences at Ukrainian Catholic University (UCU), whose collaboration allowed to create SoftServe Research Group at UCU. The authors thank Rostyslav Hryniv for helpful and valuable feedback.

% ---- Bibliography ----
%
% BibTeX users should specify bibliography style 'splncs04'.
% References will then be sorted and formatted in the correct style.
%
\bibliographystyle{splncs04}
\bibliography{literature}

\appendix
\section{Appendix}

For the evaluation we use $mIoU$ (mean intersection over union), Dice score and $F1$-score.

%  IoU
The mIoU metric was used to compare our result to already existing approaches. For all images and $k$ classes is defined as follows:
$$
mIoU = \frac{1}{k} \sum\limits_{i = 1}^k \frac{TP_{ii}}{\sum\limits_{j = 1}^k FN_{ij} +\sum\limits_{j = 1}^k FP_{ij} - TP_{ii}},
$$
where $TP$, $FP$, $FN$ are numbers of true positive, false positive
and false negative pixels respectively.

% Dice
The Dice score was used to compare our results of weakly-supervised model to the first place solution with fully-supervised approach on SIIM-ACR Pneumothorax \cite{siimpneumothorax}. The metric is defined as follows:
$$
Dice = \frac{2\times TP}{(TP + FP) + (TP + FN)}
$$
where $TP$, $FP$, $FN$ are numbers of true positive, false positive
and false negative pixels respectively. If all pixels are true negative, prediction is considered correct and metric equals 1.

% F1-score
The $F1$-score was used to compare approaches on a first step step of our pipeline --- classification.
The metric is defined as follows:
$$
F1 = \frac{2\times precision \times recall}{pression + recall}
$$
where $precission=\frac{TP}{TP + FP}$, $recall=\frac{TP}{TP + FN}$, and $TP$, $FP$, $FN$ are true positive, false positive and false negative rates respectively.

\end{document}